\documentclass{article}

\usepackage[final]{arxiv}
\usepackage{graphicx}
\usepackage{epstopdf}
\usepackage{caption}

\usepackage[utf8]{inputenc} 
\usepackage[T1]{fontenc}    
\usepackage{hyperref}       
\usepackage{url}            
\usepackage{booktabs}       
\usepackage{amsfonts}       
\usepackage{nicefrac}       
\usepackage{microtype}      
\usepackage{setspace}       
\usepackage{amsmath}
\usepackage{amssymb}
\usepackage{adjustbox}
\usepackage{tablefootnote}
\usepackage[bottom]{footmisc}

\usepackage{algorithmic}
\usepackage{subcaption}
\usepackage[ruled]{algorithm2e}

\newtheorem{theorem}{Theorem}[section]

\usepackage{natbib}
\title{Feature Selection via Maximizing Distances between Class Conditional Distributions}

\author{
Chunxu CAO \And Qiang ZHANG
}

\begin{document}


\maketitle

\begin{abstract}

For many data-intensive tasks, feature selection is an important preprocessing step. However, most existing methods do not directly and intuitively explore the intrinsic discriminative information of features. We propose a novel feature selection framework based on the distance between class conditional distributions, measured by integral probability metrics (IPMs). Our framework directly explores the discriminative information of features in the sense of distributions for supervised classification. We analyze the theoretical and practical aspects of IPMs for feature selection, construct criteria based on IPMs.
We propose several variant feature selection methods of our framework based on the 1-Wasserstein distance and implement them on real datasets from different domains.
Experimental results show that our framework can outperform state-of-the-art methods in terms of classification accuracy and robustness to perturbations.

\end{abstract}

\section{Introduction}

High-dimensional datasets have many features or variables compared to observations or samples.
They are common in fields such as bioinformatics, medicine, economics, and data science, where they can capture and reveal complex phenomena.
However, they also challenge data analysis and machine learning with high cost, difficulty, complexity, and sometimes low performance. 
Therefore, it is important to use appropriate methods and techniques to handle high-dimensional datasets and extract useful information.
One of the most commonly used methods is feature selection, which selects a subset of relevant features for a given learning task.
Feature selection can facilitate data representation, reduce storage costs, and improve the efficiency and performance of learning tasks.
It can also help understand the principles of certain tasks. For example, in bioinformatics, feature selection can provide new insights from datasets.

Feature selection methods can be classified by the role of learning models: filter, wrapper and embedded methods \cite{guyon2003introduction, li2018feature, cai2018feature}. Each method has pros and cons.
Filter methods \cite{yang1999data,nie2008trace, meyer2008informationtheoretic, gu2012generalized} use predefined criteria to evaluate features by capturing the intrinsic information in the data.
They are fast and generic, but may ignore feature-algorithm interactions.
Wrapper methods \cite{kohavi1997wrappers} use the learning algorithm to evaluate features through prediction performance.
They may achieve higher accuracy than filter methods, but also limit the generalizability of the features and face the challenge of finding the optimal subset of features. This is an NP-hard combinatorial problem that requires high computational cost.
Embedded methods \cite{tibshirani1996regression,lemhadri2021lassonet} integrate the search process into the training process of a given learning algorithm.
They can balance efficiency and effectiveness, but are algorithm-specific and may vary on the same dataset.
Another limitation of learning model-based feature selection methods is the trade-off between expressiveness and computational cost.
The more expressive the model is, the more complex information it can capture, but also the higher the computational cost and the lower the interpretability.
In this paper, we focus on filter methods, for they have some advantages in terms of interpretation and computational cost due to their independence.

Filter methods select features based on intrinsic properties of the data, without involving any learning model.
The performance of filter methods depends strongly on the choice of the evaluation criterion, which measures the influence of the features.
The optimal feature subset is obtained by optimizing this criterion according to a search strategy.
Many evaluation criteria have been proposed for feature selection. Some methods are based on the dependence between features and the target variable \cite{hall1999feature, song2012feature}, and aim to find the feature set which is the most relevant to the labels.
Some methods are based on information theory, and explore the conditional information contained in features \cite{yang1997comparative, fleuret2004fast, hanchuanpeng2005feature, meyer2008informationtheoretic, brown2012conditional}.
Most of them have good theoretical properties and practical performance.

However, there are still some drawbacks of existing criteria, most of them used in supervised classification tasks do not directly reflect the discriminative ability of features, but rather the dependency or correlation to the label.
In dependency maximization methods, the goal is to find the most relevant features to labels, not the most discriminative ones, and the relevance or dependency measure is predefined by humans.
Moreover, the detection ability of criteria is limited by their assumptions, e.g., Pearson's correlation only detects linear relationships.
Some previous works alleviate this problem by nonlinear transformations or kernel tricks.
Kernel-based dependency estimation, such as HSIC can capture nonlinear and complex associations, but it depends heavily on the choice of kernel \cite{song2012feature}.
Challenges for kernel-based criteria also include scale and monotone transform \cite{ding2017robustequitable}.
For information theoretical methods, relevance is represented by some information measures, but the goal is still not to directly find features with the most discriminative power.
Moreover, estimating mutual information is difficult and computationally expensive, as the joint probability density of feature and target variable may be high-dimensional and complex. There is still a lack of attention to develop a criterion which is directly related to the discriminative ability of features between classes. Some people have noticed this need, but to the best of our knowledge, no systematic framework has been proposed yet.

In this paper, we propose a distance-maximizing feature selection framework based on integral probability metrics between class conditional distributions, which is suitable for multi-class supervised feature selection problems, and is directly related to the discriminative power.
Our framework is intuitive and natural, as it assumes that good features should be discriminate between different classes in a probabilistic sense.
Our framework is also flexible, as we can choose different measures of distance between distributions and search strategies according to the tasks.
We construct several new feature selection methods based on our framework to demonstrate its flexibility.
We evaluate these methods on several real-world benchmark datasets and compare them with popular feature selection methods.
The experimental results show that our methods are the best or among the best performers, and are applicable to a wide range of problems.

The main contributions of our work include the following:

\begin{itemize}
	
	\item We introduce a novel feature selection framework, which utilize integral probability metrics to measure the distance between class conditional distributions, and selects features that maximize this quantity. 
	\item We show the natural connection between our framework and supervised learning problems, and demonstrate the theoretical benefits of our criteria.
	\item We propose a variant of our framework based on the 1-Wasserstein distance, and demonstrate the theoretical justification, estimation, and derivative feature selection algorithms.
	\item We implement experiments on real-world benchmark datasets and show the effectiveness of our methods.
\end{itemize}

The rest of the paper is organized as follows.
In Section \ref{Distributional_Distance_based_Feature_Importance}, we  introduce the distance-maximization framework.
We present the natural connection between our framework and supervised learning problems, and show that the criteria in our framework are expressive, easy to estimate, and concentrated.
Section \ref{Wasserstein_Distance_based_Feature_Selection} presents a feasible variant of our framework using the 1-Wasserstein distance.
We demonstrate the properties of such a measure that makes it advantageous to be used in feature selection.
We show the exact unbiased estimation of the 1-Wasserstein distance in the 1-dimensional case, and an efficient approximate empirical estimator in the high-dimensional case, with convergence results in the probability of the empirical measures.
We also provide several feature selection algorithms based on 1-Wasserstein distance as demos of our framework.
In Section \ref{Experiments}, we report our experiments on benchmark datasets from various domains, including speech recognition, image classification, bioinformatics and sensor data.
Finally, we conclude this paper in Section \ref{Conclusion}.

\section{Distributional Distance based Feature Importance}
\label{Distributional_Distance_based_Feature_Importance}

We begin by describing the supervised feature selection problem, which is the focus of this paper.
We then present our feature selection framework for solving this problem in Section \ref{IPMs_as_criteria} and Section \ref{seleaction_strategy}, introducing our criteria in the former and selection strategies in the latter.

\subsection{Supervised Feature Selection Problem}

Consider a data-generating model $p(\mathbf{x}, y)$ over a $d$-dimensional space, where $\mathbf{x} \in \mathcal{X} \subset R^d$ is the covariate and $y \in \mathcal{Y}={\{c_1,\dots,c_K\} }$ is the response, such as class labels.
The goal of supervised learning is to find the best function $f^\star(\mathbf{x}) $ from the function class $\mathcal{F}$ for predicting ${y}$.
To measure the goodness of the prediction, the loss function $\mathcal{L}$ is used to quantify the difference between prediction $f(\mathbf{x})$ and the ground truth $y$.
We want to find a function $f^\star(\mathbf{x}) $ that achieves the minimum risk $R_{exp}(f)$, which is the loss $\mathcal{L}$ in expectation sense, among all the candidate functions in $\mathcal{F}$.
\begin{equation}
f^{\star} =\underset{f \in \mathcal{F}}{\arg\min} R_{exp}(f) = \underset{f \in \mathcal{F}}{\arg\min} E[\mathcal{L}(f(\mathbf{x}), y)]
\label{supervised_learning_problem}
\end{equation}

Feature selection is to find the feature subset $\mathcal{T}^\star$ with biggest utility $U$ from the original feature set $\mathcal{S}$ that achieves the minimum risk $R_{exp}(f)$ among all the candidate feature subset $\mathcal{T}$ with required size $m$.
\begin{equation}
\mathcal{T}^\star = \underset{ \mathcal{T} \subset \mathcal{S} ,f \in \mathcal{F}}{\arg\max} U(\mathcal{T}), \quad |\mathcal{T}|=m
\label{feature_selection_problem}
\end{equation}
where the $|\mathcal{T}|$ is the cardinality of the set $\mathcal{T}$.
In the supervised learning case, it's to find the feature subset $\mathcal{T}^\star$, on which we could find a best function $f^\star(\mathbf{x}_{\mathcal{T}}) $ from $\mathcal{F}$ to achieve the minimum risk $R(f)$ in expectation sense among all the candidate feature subset.
\begin{equation}
	\begin{aligned}
	\mathcal{T}^\star & = \underset{ \mathcal{T} \subset \mathcal{S} ,f \in \mathcal{F}}{\arg\max} U(\mathcal{T}) \\ & = \underset{ \mathcal{T} \subset \mathcal{S} ,f \in \mathcal{F}}{\arg\min} R_{exp}( f(\mathbf{x}_{\mathcal{T}}) ) \\ & = \underset{ \mathcal{T} \subset \mathcal{S} ,f \in \mathcal{F}}{\arg\min} E[\mathcal{L}(f(\mathbf{x}_{\mathcal{T}}), y)], \quad |\mathcal{T}|=m
	\label{supvised_feature_selection_problem}
	\end{aligned}
\end{equation}

However, we often need to evaluate features without training the classifier in practice, so we need to choose appropriate quantification function $U(\mathcal{\cdot})$ and a selection strategy independent of learning models to find the optimal subset $\mathcal{T}^\star$.

\subsection{IPMs as the Criteria of Feature Selection}
\label{IPMs_as_criteria}

Measuring the utility of a feature is a fundamental problem in statistics and machine learning, where a number of criteria are potential candidates for feature selection.
The difficulty arises from the fact that the relationship between covariate and response may be sophisticated, while we often have limited samples with no prior knowledge.
We could alleviate this problem in supervised classification tasks, where important features should be discriminative in different classes.
In the supervised learning case, we can find a subset of features that are actually the most discriminative among the candidates in expectation, so the utility $U$ could be represented by the discriminative power among the labels.
In this setting, we could get rid of the need to decide which mode of the relationship between the covariate and the response should be taken into account.
Assuming that we are given samples$(\mathbf{X},\mathbf{y})$, where $\mathbf{X}$ is a data matrix corresponding to $K$ classes and $\mathbf{y}$ are the associated labels, the feature selection problem can be reformulated as follows:
\begin{equation}
	\mathcal{T}^{\star} =\arg \max _{\mathcal{T} \subseteq \mathcal{S}} \sum_{i=1}^K\sum_{j=1}^K F_d(X_{\mathcal{T}}^i,X_{\mathcal{T}}^j),\text { s.t. }|\mathcal{T}|=m
	\label{prob_dist_problem}
\end{equation}
where $F_d(\cdot,\cdot)$ is the discriminant function that estimates the discriminative power of features based on data from different classes corresponding to the features, $X_{\mathcal{T}}^i,X_{\mathcal{T}}^j$ are the samples of the $i$-th and $j$-th class corresponding to the feature subset $\mathcal{T}$ respectively.

However, although we can leverage the label information, to find a good criterion that satisfies the feature selection requirements is still challenging.
It's natural to expect the criteria to be general, so that we can perform feature selection in different domains and detect all the mode of distinctive behavior of features when the label changes.
Convergence and consistency are also necessary to hope that we can get close to the truth and that the selected features perform well on test data.
Moreover, considering that feature selection has been widely applied in knowledge discovery, it would be appreciated if the criterion has good theoretical justification meanwhile intuitive interpretation.

To meet these requirements, we propose a feature selection framework that maximizes the distance between class conditional distributions, in which we construct criteria based on the distance between class conditional distributions measured by integral probability metrics (IPMs) $\gamma_{\mathcal{F}}(P,Q)$, to meet these demands in feature selection.
\begin{equation}
\mathcal{T}^{\star} = \arg\max_{\mathcal{T}\subset S} \sum_{i}^{K}\sum_{j}^{K} \gamma_{\mathcal{F}} ( p^{\prime}(\mathbf{X}_{\mathcal{T}}|y=c_i) , p^{\prime}(\mathbf{X}_{\mathcal{T}}|y=c_j) |)
\label{IPM_framework}
\end{equation}
\begin{equation}
	\gamma_{\mathcal{F}}(P,Q): =\underset{f\in\mathcal{F}}{sup}| \int _{M}fdP- \int _{M}fdQ |
	\label{definition_of_IPMs}
\end{equation}

where $\mathcal{F}$ is a function class contains real-valued bounded measurable functions defined on measurable space $M$.

There are many benefits of utilizing IPMs in feature selection.
IPMs are metrics, which have good theoretical justification and interpretation, they take into account all the difference between probability measures, which guarantee the generality and versatility to the criteria.
The intrinsic connection of IPMs to the notion of weak convergence of probability measure makes IPMs meet the demand of convergence on empirical measures in feature selection straightforwardly.
Besides, IPMs measure the distance between class conditional distributions in classification problems, the associated optimal loss function bounds the performance and relates to the complexity of learning models for a given task.
In addition to those beneficial characteristics of IPMs, estimating them are not difficult, many works studied the estimation of IPMs, by which we can compute the criteria easily with confidence.
Hence, thus it would be effective and feasible to apply IPMs in feature selection in supervised learning.

We will demonstrate the derivation and benefits of using the distance between class conditional distribution measured by IPMs as the feature selection criteria in detail in following.

\subsubsection{Distance of Class Conditional Distribution in Supervised Learning}
The core of designing the feature selection method is to make clear on which the criteria measures, we measure the distance between class conditional distribution to merit features for its predominant role in supervised learning.

To demonstrate it in detail, we need to revisit the supervised learning problem \ref{supervised_learning_problem}.
Suppose that we care about accuracy, that the loss function $\mathcal{L}$ is 0-1 loss, the risk $R_{exp}(f)$ is
\begin{equation*}
R_{exp}(f) = E [\sum_{k=1}^{K}[\mathcal{L}(y=c_k, f(\mathbf{x})) ] p(y=c_k|\mathbf{x})]
\label{expected_risk}
\end{equation*}

To minimize the expected risk, the optimal classifier function $f^\star$ should minimize the loss with respect to each input, leading to the Bayesian optimal classifier, i.e., classifying samples to labels with maximum posterior probability:
$$f(x)= \underset{c_k\in\mathcal{Y}}{\arg\max} P(c_{k}|X=x)$$
In this case, the classifier $f^\star(\cdot)$ could reach the minimum risk $R^{\star}$ in a probability sense, which limits the performance of the classifier in this task.
Prediction could be straightforward if we could estimate the posterior probability $p(y|\mathbf{x})$ by the class conditional distribution $p(\mathbf{x}|y)$ and the class prior probability $p(y)$.
However, the class prior $p(y)$ is unknown, and in practice it is treated as appropriate, the core of performance in supervised learning is usually the class conditional distribution $p(\mathbf{x}|y)$.
Thus, it's natural to study the discriminant behavior of features by exploring the class conditional distributions.

More specifically, we want the class conditional probability distributions between classes to be far apart.
To illustrate this, we consider the case of using a linear classifier in a 1-D binary classification problem.
In such a case, the classifier we find should divide the feature space into two parts ${\Omega}_1,{\Omega}_2$ corresponding to the label $+1,-1$, minimizing the 0-1 loss could be equivalent to minimize the probability of misclassification $\mathbf{P}_e$.
\begin{equation}
	\mathbf{P}_{e}=P(\mathbf{x}\in {\Omega_1},y=-1) + P(\mathbf{x}\in {\Omega_2},y=+1)
	\label{binary_error_prob}
\end{equation}
The probability of error classification could be interpreted as the overlap between two class conditional distributions.
Intuitively, if we can reduce the overlap between different class conditional distributions, we can reduce the probability of error classification.
Otherwise, if the two distributions overlap completely, we can not separate anyway.
Moreover, if there is no overlap between class conditional distributions, the distance is the upper bound of the performance of classifier.
The discriminative power of the feature is now represented by the distance between the class conditional distributions.

There are also many rigorous works studying the relationship between distance in class conditional distributions and binary classification problem.
\cite{nguyen2009surrogate} demonstrated the correspondence that for any loss function $\mathcal{L}$, there exists a corresponding $\phi$-divergence such that the negative $\phi$-divergence between class conditional distributions equals to the optimal risk $R_{\mathcal{F}}^{\mathcal{L}}$, which is associated with a binary classifier $f\in\mathcal{F}$ that classifies the class conditional distributions $P$ and $Q$ using the loss function $\mathcal{L}$.
$$\begin{aligned} R_{\mathcal{F}}^{\mathcal{L}}= & \inf _{f \in \mathcal{F}} \int_{\mathcal{X}} \mathcal{L}(y, f(\mathbf{x})) d p(\mathbf{x}, y) \\ = & \inf _{f \in \mathcal{F}}\left\{p(y=+1) \int_{\mathcal{X}} \mathcal{L}(+1, f(x)) d p(\mathbf{x}|y=+1)\right. \\ & \left.+ p(y=-1) \int_{\mathcal{X}} \mathcal{L}(-1,f(x)) d p(\mathbf{x}|y=-1)\right\}\end{aligned}$$

The $\phi$-divergence is defined based on convex function $\phi:[0,\infty)\to (-\infty,\infty]$ and measurable space $M$:
\begin{equation}
	D_{\phi}(P,Q):= \left\{ \begin{matrix} \int _{M}\phi(\frac{dP}{dQ})dQ, \quad P<Q \\ + \infty , \quad otherwise \\ \end{matrix} \right.
	\label{phi_divergence}
\end{equation}
where $P$ and $Q$ are probability measures.
Several well-known divergences could be obtained by choosing the $\phi$, e.g., the Kullback-Liebler (KL) divergence corresponding to the $\phi(t) = tlogt$, the Hellinger distance could be obtained by setting the $\phi(t)= (\sqrt{t} -1)^2$, and they relate to the loss function $\mathcal{L}(1,\alpha)=-\frac{\alpha}{\epsilon}, \mathcal{L}(-1,\alpha) = \frac{e^{\alpha -1}}{1-\epsilon}$ and the exponential loss, respectively.

\cite{sriperumbudur2009integral} studied the relationship between IPMs and binary classification problem.
They showed that the IPMs between the class conditional distribution $P$ and $Q$ is the negative of the optimal risk $R_{\mathcal{F}}^{\mathcal{L}}$ associated with a binary classifier.
They also pointed out that the 1-Wasserstein distance and Dudley distance relate to the 1-Lipschitz classifiers and the bounded 1-Lipschitz classifiers and bound the margins of them, respectively.
Therefore, the distances between the class conditional distribution determine the upper bound of the performance corresponding to the classifier which is restricted in certain function class.
If the class conditional distribution is complex and overlapping, the supervised learning problem might be hard to classify, even worse, inseparable, if the class conditional is distinct and simple, the problem may be easy to solve.

Therefore, we focus our attention on the distance between the class conditional distributions to study the utility and behavior of the feature set $\mathcal{T}$.

\subsubsection{Feature Selection Maximizing the Distance between Class Conditional Distributions}

As the goal of feature selection in supervised learning is to select a useful subset $\mathcal{T}^\star$ from the original feature set $\mathcal{S}$ which preserves the useful information contained in $\mathcal{S}$ to maintain or even improve the performance of classifiers, it might be helpful to find a feature subset that shows the largest distance between the class conditional distributions.
This idea is intuitive, a good feature should vary significantly when the label changes, the lager the discrepancies the feature shows, the more discriminative the feature is.
Thus, the optimal feature subset $\mathcal{T}^\star$ is the one that shows the largest distance $D^\star$ across the class conditional distributions, the feature selection in the binary classification problem can then be formulated as follows:
\begin{equation}
	\mathcal{T}^{\star} = \arg\max_{\mathcal{T}\subset S}  {D} ( p(\mathbf{x}_{\mathcal{T}}|y=+1), ( p(\mathbf{x}_{\mathcal{T}}|y=-1) |),\quad |\mathcal{T}|=m
	\label{feature_distance_maximization_binary}
\end{equation}
where the ${D}$ is the distance measure between the distributions, it estimates the discriminative power of the feature set, plays the role of the function $F_d$ in formulation \ref{prob_dist_problem}.

In practice we are often given data matrix $\mathbf{X}$ and a label vector $\mathbf{y}$ consisting of samples drawn from the generation distribution, we need to perform inference based on the empirical class conditional distribution corresponding to the feature set $\mathcal{T}$.
To find the optimal feature subset $\mathcal{T}^{\star}$ is actually to find the submatrix $\mathbf{X}_{\mathcal{T}^\star}$ that has the largest distance over the empirical class conditional distribution $p^\prime(\mathbf{X}_{\mathcal{T}} |y )$ among all the feature subset $\mathcal{T}$ of size $m$.
\begin{equation}
	\mathcal{T}^{\star} = \arg\max_{\mathcal{T}\subset S}  {D} ( p^{\prime}(\mathbf{X}_{\mathcal{T}}|y_+), p^{\prime}(\mathbf{X}_{\mathcal{T}}|y_-) |),\quad |\mathcal{T}|=m
	\label{feature_distance_maximization_binary_emprical}
\end{equation}
It's natural to extend this idea to multi-class supervised learning problems.
We can transform the original feature selection problem \ref{supvised_feature_selection_problem} into maximizing the distances in all pairs of empirical class conditional distributions:
\begin{equation}
	\mathcal{T}^{\star} = \arg\max_{\mathcal{T}\subset S} \sum_{i}^{K}\sum_{j}^{K} {D} ( p^{\prime}(\mathbf{X}_{\mathcal{T}}|y=c_i) , p^{\prime}(\mathbf{X}_{\mathcal{T}}|y=c_j) |)
	\label{feature_expect_discrepancy_maximization}
\end{equation}

The core of finding criteria for the feature selection problem now becomes choosing an appropriate measure to quantify the difference or distance between class conditional distributions.
In the case of supervised learning, it's natural to consider IPMs as the measure for their dominant role in binary classification problem.
Furthermore, since we are concerned with the difference classifier $f$ shown under class conditional distribution, we can reformulate \ref{feature_expect_discrepancy_maximization} to consider the dual norm for the difference between class conditional distributions on symmetric convex set of measurable functions $f\in \mathcal{F}$:
\begin{equation}
|| \alpha ||_{\mathcal{F}} = \underset{f\in \mathcal{F}}{max} \left\{ \int _{x}f(x)d \alpha(x):f \in \mathcal{F} \right\} .
\end{equation}
where the $\alpha = P-Q$, it's the definition of IPMs \ref{definition_of_IPMs}, it's consistent with the feature selection in supervised learning.

These properties of IPM show that the choice of IPMs to measure the distance between probability measures is considerable.
Now we state the main result of this paper, that the utility of feature set $U(\mathcal{T})$ in supervised learning could be quantified by criteria based on IPMs between the empirical class conditional distributions restricted to $\mathcal{T}$.
In the binary case, the quantified utility of the feature set $\mathcal{T}$ is 
\begin{equation}
	U(\mathcal{T}) = \gamma_\mathcal{F}( p^\prime(\mathbf{X}_{\mathcal{T}}|y=+1), p^\prime(\mathbf{X}_{\mathcal{T}}|y=-1))
	\label{binary_case_ipm_imp}
\end{equation}
In case of multiple classes, we can obtain a distance matrix $\mathbf{D}^\mathcal{T}$ whose dimension is the number of classes, entry $\mathbf{D}_{ij}^\mathcal{T}$ is the distance between class conditional probability measures of the class $i,j$ corresponding to feature subset $\mathcal{T}$.
This matrix contains all the distributional disparity behavior of a particular feature subset, we can summarize the matrix efficiently by the Frobenius norm, for the invariant property and the good physical sense.
This quantity is zero if and only if there exists no difference for feature set $\mathcal{T}$ among all classes, it summarizes the magnitude of disparities of class conditional distributions corresponding to feature set $\mathcal{T}$.
We use this quantity as the quantifying function of feature utility, and introduce the framework that takes this quantity as the criterion.
\begin{equation}
	U(\mathcal{T}) = ||\mathbf{D}^\mathcal{T}||_{Frob}^2 = \sum_{i=1}^K\sum_{ j=1 }^K(\mathbf{D}_{ij}^\mathcal{T})^2 = \mathbf{tr}\mathbf{D}^\mathcal{T} (\mathbf{D}^\mathcal{T})^T
\label{criteria_multi_class}
\end{equation}
where the $(\mathbf{D}^\mathcal{T})^T$ is the transpose of $\mathbf{D}^{\mathcal{T}}$ and $\mathbf{D}_{ij}^\mathcal{T}=IPM( p^\prime(\mathbf{X}_{\mathcal{T}}|y=c_i), p^\prime(\mathbf{X}_{\mathcal{T}}|y=c_j))$.

Therefore, if we want the bound of performance of classifier is higher while the classifier is simpler, feature selection in supervised learning problem is actually to find features that make samples from different classes scatter as far as possible.

We must say that feature evaluation using the distance or divergence between class conditional probability of feature is not a novel idea.
About two decades ago, a Kullback J-divergence based feature selection method has been proposed\cite{novovicova1996divergence}, the author propose a feature selection method suitable for binary classification of multimodal data, based on the Kullback J-divergence between class conditional probability.
In recent years, some researchers have proposed distance-based feature selection methods, but they neither propose a systematic and versatile framework nor apply the methods to multi-class feature selection problems.
To our knowledge, we are the first to derive IPMs distance maximization framework for feature selection in multi-class supervised learning problem.

\subsubsection{IPM Criteria is Expressive}

In feature selection, the criteria should be expressive enough to capture the relevance and redundancy of features for certain task, and to detect the functional relationship between input and labels.
For instance, Pearson's correlation could only detect the linear relationship between variables, which leads to a limited range of application.
While the HSIC, a kernel-enhanced correlation measure, is a good criterion for its ability to detect the desired functional dependence \cite{song2012feature}.

In our feature selection framework, expressiveness of criteria means the ability to measure the difference between class conditional distributions.
IPMs are not the only way to compare the difference between probability measures, but they have some advantages in quantifying the distance between distributions when compared to other difference measures, such like $\phi$-divergence.
The most important one is that IPMs are metrics, feature selection criteria based IPMs could be expressive and consistent in different tasks due to the metric nature of IPMs:
\begin{itemize}
	\item $\gamma_\mathcal{F}(P, P)$ = 0 for any probability distribution $P$.
	\item $\gamma_\mathcal{F}(P, Q)$ > 0 for any two different probability distributions $P$ and $Q$.
	\item $\gamma_\mathcal{F}(P, Q)$ = $\gamma_\mathcal{F}(Q, P)$ for any two probability distributions $P$ and $Q$.
\end{itemize}
Note that the Wasserstein distance and the Dudley metric are metrics, but the MMD is not a metric unless the RKHS is universal.

IPMs are defined as the supremum of the difference between the expectations of a function under two distributions, where the function belongs to a certain function class.
The distance measured by IPMs between two probability distributions is zero if and only if they are the same one, regardless of the choice of IPM \cite{dudley2008real}.
If these two distributions are the same one, then their expectations are equal for every function in the class, and the supremum of the difference is zero.
Conversely, if they are different, then there exists at least one function in the class such that their expectations are different, and the supremum of the difference is positive.
This holds for any choice of function class in IPM, as long as it is non-trivial (i.e., not a constant function).
Thus, IPMs could detect all modes of difference between class conditional distributions.

IPMs have many other important advantages that make them suitable for performing feature selection in various domains.
For example, they can handle cases where distributions have no common support or even the same dimension, and they can be insensitive to mismatches in the support of distributions, which makes IPMs a stronger and more consistent notion than many other measures for comparing the difference in a pair of distributions.
Furthermore, there is a natural relationship between IPMs and binary classification, which makes it consistent to select the feature subset $\mathcal{T}$ that maximizes the IPMs for the subsequent task.
Moreover, IPMs treat the probability distributions directly, which guarantees their expressiveness in various domains.
For datasets collected in practice, which are usually digital recordings, we can always extract the empirical measures from the data, regardless of the type of data.
There are many applications of IPMs in two-sample test and generative models, covering a wide range of areas in the literature, which also support the expressiveness and versatility of IPMs \cite{rubner2000earth, dudley2008real, song2012feature}.

IPMs are distance measures that depend on the function class used to compare the expectations of two distributions. Different function classes endow IPMs with different abilities to capture the intrinsic information of class conditional distributions.
The Wasserstein distance uses Lipschitz functions and reflects the geometry of distributions. It can handle complex problems such as comparing distributions from different supports or models. It is suitable for feature selection when we care about the geometric structure of the sample space.
The Dudley metric uses bounded continuous functions and metrizes the weak topology. It measures the worst case difference between two distributions over all bounded continuous functions. It is suitable for feature selection when we are interested in the weak topology on the set of all probability measures on a given metric space. The closeness of two probability measures under the Dudley metric implies their agreement on all weakly continuous functions.
The MMD uses an RKHS of functions and is sensitive to the geometry of the feature space induced by a kernel. It measures the maximum difference between two distributions over all functions in the RKHS. It can examine a particular type of dependence using a kernel function. It is suitable for feature selection when we want to exploit the kernel trick and achieve a fast rate of convergence to the population value that is independent of the dimension of the space.
The TV distance uses indicator functions and is sensitive to the mass or density of distributions. It detects any difference between two distributions regardless of their structure or support. It is suitable for feature selection when we want to use a simple and robust measure that does not require any parameter tuning or prior knowledge.

We note that all IPMs can be used in our feature selection framework, and they incorporate different prior knowledge into the feature utility estimation. In the following sections, we present our framework and its theoretical and empirical results.

\subsubsection{Estimation of IPMs}

One of the reason to employ IPMs in feature selection is that estimating a IPM is not difficult, which is beneficial in practice.

The empirical estimation could be obtained from the definition of IPMs.
Given data sets $\mathbf{X}^{(1)} = \{\mathbf{x}_1^{(1)},\dots,\mathbf{x}_m^{(1)}\}, \mathbf{X}^{(2)} = \{\mathbf{x}_1^{(2)},\dots,\mathbf{x}_n^{(2)}\} $ consist of i.i.d. samples drawn from $P$ and $Q$ respectively, the estimation of certain IPM $\gamma_{\mathcal{F}}$ corresponding to function class $\mathcal{F}$ is
\begin{equation}
\gamma_{\mathcal{F}}(P_{m},Q_{n}) = \underset{f\in\mathcal{F}}{sup}| \sum _{i=1}^{N}\tilde{Y}_{i}f(\mathbf{x}_{i})|
\label{empirical_estimation_of_IPMs}
\end{equation}

where $P_m, Q_n$ are the empirical measures corresponding to $P$ and $Q$, $N=m+n,\tilde{Y}_{i}=\frac{1}{m}$ if $\mathbf{x}_i = \mathbf{x}^{(1)}_i$, otherwise $\tilde{Y}_{i}= -\frac{1}{n}$.

Since IPMs correspond to certain function classes, the estimation could be more straightforward if restricted to certain function classes.
\cite{sriperumbudur2012empirical} showed the estimation of various distances on empirical measures and their estimators are easy to compute, including the 1-Wasserstein distance, Dudley metric, total variation(TV) distance, and maximum mean discrepancy(MMD).
They also pointed out that the 1-Wasserstein distance, Dudley metric, and TV could be obtained by solving linear programs, while the MMD could be obtained in closed form.
Here we present the empirical estimator of the Wasserstein distance and the closed-form estimator of the MMD for completeness, we recommend interested readers to refer to \cite{sriperumbudur2012empirical} for more details.
\begin{theorem}[Estimator of the Wasserstein distance \cite{sriperumbudur2012empirical}]
The following function solves \ref{empirical_estimation_of_IPMs} for 1-Lipschitz function class $\mathcal{F}_{W}:= \left\{ f:||f||_{L}\leq 1\right\}$,
\begin{equation}
	f_\alpha(x):=\alpha \min _{i=1, \ldots, N}\left(a_i^{\star}+\rho\left(x, X_i\right)\right)+(1-\alpha) \max _{i=1, \ldots, N}\left(a_i^{\star}-\rho\left(x, X_i\right)\right)
\label{empirical_estimator_func_of_mmd}
\end{equation}
now the estimation of 1-Wasserstein distance is:
\begin{equation}
W(P_{m},Q_{n})= \sum _{i=1}^{N}\tilde{Y_{i}}a_{i}^{*}.
\label{empirical_estimator_of_mmd}
\end{equation} 
where the $a_{i}^{*}$ is the solution of following linear program
\begin{equation}
	\max _{a_1, \ldots, a_N}\left\{\sum_{i=1}^N \widetilde{Y}_i a_i:-\rho\left(X_i, X_j\right) \leq a_i-a_j \leq \rho\left(X_i, X_j\right), \forall i, j\right\}
\label{empirical_estimator_solution_of_mmd}
\end{equation}

\end{theorem}

Since MMD is a kernel-induced IPM, it shares the kernel's advantage that the computational complexity is independent of the dimension of the distribution, i.e. it's free from the curse of dimension.
\begin{theorem}[Estimator of MMD \cite{sriperumbudur2012empirical}]
	Given a strictly positive definite kernel $k$, then the following function solves \ref{empirical_estimation_of_IPMs} for corresponding reproducing kernel Hilbert space(RKHS) $\mathcal{H}$, $\mathcal{F}_{k}:= \left\{ f:||f||_{\mathcal{H}}\leq 1\right\}$:
	
	\begin{equation}
f=\frac{1}{\left\|\sum_{i=1}^N \widetilde{Y}_i k\left(\cdot, X_i\right)\right\|_{\mathcal{H}}} \sum_{i=1}^N \widetilde{Y}_i k\left(\cdot, X_i\right)
		\label{empirical_estimator_func_of_1_wsd}
	\end{equation}
	now the estimation of MMD is:
	\begin{equation}
\gamma_k\left({P}_m, {Q}_n\right)=\left\|\sum_{i=1}^N \tilde{Y}_i k\left(\cdot, X_i\right)\right\|_{\mathcal{H}}=\sqrt{\sum_{i, j=1}^N \tilde{Y}_i \tilde{Y}_j k\left(X_i, X_j\right)}.
		\label{empirical_estimator_of_1_wsd}
	\end{equation} 
	
\end{theorem}

Since the utility of features based on IPMs \ref{criteria_multi_class} is the sum of the squares of the empirical estimates of the IPMs, our criteria could be easily estimated based on the estimates of the IPMs.

Note that different choices of IPMs also imply different computational costs.
The Wasserstein distance and the MMD have scalable algorithms based on approximate estimators and kernel methods respectively, but the Dudley metric is more challenging to compute and requires approximation methods.
Moreover, the Wasserstein distance and the MMD are consistent under mild conditions on the function class and the underlying space, while the Dudley metric requires stronger conditions on the function class for consistency.

\subsubsection{IPM Criteria is concentrated}
\label{convergence_of_ipms}

In practice, we are usually given a set of samples for feature selection with the expectation that the selected features will be reliable and remain effective in the test data, so we need the criteria to concentrate to the population values.
The core of our criteria is the convergence of IPMs which has been studied extensively, that IPMs are concentrated under mild conditions \cite{geer2000empirical}.
We follow the terminology and notation from \cite{sriperumbudur2012empirical}.
Given a probability measure $Q$, define the $L_r$ norm of the function $||f||_{Q,r}:=(\int |f|^r dQ)^{1/r},~r\le1$ and denote the metric space induced by this norm as $L_r(Q)$, then denote the covering number of $L_r(Q)$ as $N(\varepsilon ,\mathcal{F},L_{r}(Q))$, and denote the logarithm of this covering number as the entropy of the function class $\mathcal{F}$ using the metric $L_r(Q)$.
Before presenting the general result on the consistency of IPMs shown in \cite{sriperumbudur2009integral}, we define the minimal covering function $F(x):=sup_{f\in\mathcal{F}}|f(x)|$ associated with the function class $\mathcal{F}$.
\begin{theorem}
Suppose the following conditions hold:
\begin{enumerate}
	\item $  \int _{S}FdP<\infty $
	\item $  \int _{S}FdQ<\infty $
	\item $ \forall \varepsilon >0, ~\frac{1}{m}H(\varepsilon ,\mathcal{F},L^{1}(P_{m}))\rightarrow 0~as~m \rightarrow \infty .$
	\item $ \forall \varepsilon >0, ~\frac{1}{n}H(\varepsilon ,\mathcal{F},L^{1}(Q_{n}))\rightarrow 0~as~n \rightarrow \infty .$
\end{enumerate}
Then, $| \gamma_{\mathcal{F}}(P_{m},Q_{n})- \gamma_{\mathcal{F}}(P,Q)| \overset{a.s.}{\rightarrow}0 ~as ~m,n \rightarrow \infty .$
\end{theorem}
Therefore, the estimation of IPMs between empirical class conditional distribution close to its counterpart on the population for features, which leads to the convergence of our criteria.

Due to the core distinction among IPMs is the choice of function class, certain convergence behavior should be considered with respect to specific IPM.
\cite{sriperumbudur2009integral} showed that the estimates of the IPMs(especially the Wasserstein distance, the Dudley metric, and the MMD) are strongly consistent and exhibit good rates of convergence.
They also show that although the empirical estimator of the TV distance is not strongly consistent, it could be bounded by strongly consistent IPMs.
Moreover, under certain boundedness and smoothness assumptions, it has been shown that the MMD converges at a rate of $O(n^{-1/2})$, where $n$ is the sample size \cite{sriperumbudur2009integral}.
This rate is faster than the linear programming based estimators of the Wasserstein distance and the Dudley metric when the dimension is larger than $2$, and is independent of the dimension, which means that MMD may be able to handle high-dimensional datasets.

Note that under certain specific assumptions, the empirical estimator may be faster than the typical rate of linear programming based estimators.
For instance, there are many modifications of the Wasserstein distance, the computational cost of the Sliced Wasserstein distance depends on the number of random projections rather than the dimension, it could be computed in linear time in simple cases.
The Smoothed Wasserstein distance leverages Gaussian smoothing to level out local irregularities in the high-dimensional distributions and takes $O(n^{-\frac{1}{2}})$ time, it's remarkably faster than the typical $n^{-\frac{1}{d}}$ rate ($d>3$) \cite{goldfeld2020asymptotic}.

Since the IPM criteria proposed in this paper are the sum of squares of the estimates of the IPMs on empirical measures, the convergence is determined by the estimator of the IPMs, so it's also concentrated.

\subsection{Selection Strategy}
\label{seleaction_strategy}

The typical supervised feature selection problem \ref{supvised_feature_selection_problem} is a combinatorial optimization problem, that poses challenges for finding the optimal solution, i.e., selecting the feature subset with the greatest utility.

Given criteria for evaluating features, as there are $\begin{pmatrix}d \\ m \\ \end{pmatrix}$ candidates when exhaustively searching the best feature set with of size $m$, finding an optimal feature subset for problem \ref{supvised_feature_selection_problem} is NP-hard \cite{kohavi1997wrappers}, it's common to find a suboptimal solution by heuristic methods.
A widely used strategy is top-$m$, it evaluates features individually according to the criterion and ranks them, then selects the first required number of features as the best feature subset.
Such strategy transforms the original feature selection problem \ref{supvised_feature_selection_problem} as follows:
\begin{equation}
	\mathcal{T}^*=\arg \max _{\mathcal{T} \subseteq \mathcal{S}} U(\mathcal{T}) = \arg\max_{\mathcal{T} \subseteq \mathcal{S}} \sum_{{{f_i}}\in\mathcal{T}} U({{f_i}} ) , \text { s.t. }|\mathcal{T}|=m ,\quad m<d 
	\label{individually_fs_problem}
\end{equation}
The top-$m$ strategy is easy to use, but it couldn't take the interaction among features into account in the evaluation, besides, it can't tackle the redundancy in the selected features.

Another commonly used strategy is greedy search, it considers the interaction or dependency among features, and tries to avoid redundancy in the selected feature set by greedily considering features one by one.
There are many greedy search approaches, such as forward add-in selection and backward elimination \cite{guyon2002gene}.
Forward add-in selection tries to increase $U(\mathcal{T})$ as much as possible for each inclusion of features, and backward elimination goes on in verse, tries to maintain the utility of features during the removal process of features.
In general, forward add-in selection is computationally more efficient, while backward elimination provides better feature subset, for the interaction among features is considered in larger candidate subsets.
There are many existing strategies that work well in various fields, we recommend interested readers to refer to \cite{li2018feature} for a more comprehensive survey.

The range of feasible strategy is determined by criteria, for instance, if the criteria is limited to evaluate features individually, the greedy search may not applicable.
The range of feasible strategies is determined by the criteria, for instance, if the criterion is limited to evaluating features individually, the greedy search may not be applicable.
In principle, a good framework or criteria should be able to work together with a wide range of strategies. 
The distance-maximization framework proposed in this paper can be employed for feature selection with either a top-$m$, forward add-in or backward elimination, or even a mix of several strategies.

The choice of strategy should be made according to certain tasks.
Top-$m$ strategy is a good choice when feature selection is processed in knowledge discovery or the goal is just to narrow down the candidates, as this strategy, greedy search, is suitable for higher performance demand.
Mixing strategies for a better trade-off between effectiveness and efficiency is also a good idea.
Note that the greedy search approaches based on learning model-independent criteria are still filter methods, so they are still easy to interpret and versatile.

In the next section, we introduce feature selection methods based on the class conditional distribution distance maximization framework, where 1-Wasserstein distance is the measure of the distance between probability measures, top-$m$ and forward add-in are the example strategies, to show the flexibility and effectiveness of our framework, further performance improvement could be achieved by changing the strategy.

\section{Wasserstein Distance based Feature Selection}
\label{Wasserstein_Distance_based_Feature_Selection}

In this section, we construct the criteria based on 1-Wasserstein distance to measure the utility of features, as a demo of our framework.

\subsection{Wasserstein Distance based Criteria}
1-Wasserstein is a special case of p-Wasserstein distance, which measures the distance between two probability measures by the optimal cost of transporting one measure to the other.
The Wasserstein distance between probability measures $P$ and $Q$ on $\mathcal{X}$ is defined as follows
\begin{equation}
W_{p}(P ,Q)= (~ \underset{\gamma \in \Gamma(P ,Q)} {\inf}  \int _{\chi \times \chi}||x-y||^{p}d \gamma(x,y) ~)^{1/p}
\label{p_wsd_definition}
\end{equation}
where $\Gamma(P ,Q)$ is the set of joint probability measures $\gamma$ on $\mathcal{X} \times \mathcal{X}$ satisfying the condition that the marginal probability distributions of the elements $\gamma$ are $P$ and $Q$, respectively.

The Wasserstein distance has a nice interpretation in terms of mass transport, where it can be seen as the minimum cost of moving from one distribution to another according to a given cost function.
This physical meaning of Wasserstein distance is natural to be applied in feature selection problem, as the discrepancy of certain feature shows between different class is the distance between class conditional probability density, the Wasserstein distance here means the minimum effort needed to distort the distribution of feature corresponding to the change of labels, which coincide to the supervised classification problem and the probabilistic modeling of data.
Therefore, constructing criteria based on Wasserstein distance is a good choice for feature selection.

If we set the $p=1$, it's the 1-Wasserstein distance, for convenience, we denote it by $W_1$
\begin{equation}
W_1(P ,Q)= \underset{\gamma \in \Gamma(P ,Q)}{\inf}
\int _{\chi \times \chi} |x-y| d \gamma(x,y)
\label{1_wsd_definition_prob}
\end{equation}
we can also define $W_1$ distance like the IPMs \ref{definition_of_IPMs}
\begin{equation}
W_1(P,Q) = \underset{f\in \mathcal{F}_w}{sup}| \int_{\mathcal{X}} f dP -  \int_{\mathcal{X}} f dQ|
\label{1_wsd_definition_ipm}
\end{equation}
where $\mathcal{F}_w$ is the set of all 1-Lipschitz functions on $\mathcal{X}$.

As a typical integral probability metric, the $W_1$ distance relates to the margins of 1-Lipschitz classifiers in binary classification; the smaller the $W_1$ distance between class conditional distributions, the less smooth and more complex the resulting Lipschitz classifier is \cite{sriperumbudur2012empirical}.
It has also been shown that the $W_1$ distance between class conditional distributions is the negative of the optimal risk associated with 1-Lipschitz classifiers \cite{sriperumbudur2009integral}.
It's also usually more robust than setting $p$ greater than 1 \cite{peyre2019computational}.
We can evaluate the utility of features using the $W_1$ distance, in the binary classification problem the utility of the feature set $\mathcal{T}$ is
\begin{equation}
	U(\mathcal{T}) = W ( p^\prime(\mathbf{X}_{\mathcal{T}}|y_{+}) , p^\prime(\mathbf{X}_{\mathcal{T}}|y_{-}) )
	\label{binary_wsd1_imp}
\end{equation}
in multi-class case, it is
\begin{equation}
	U(\mathcal{T}) = ||\mathbf{D}^\mathcal{T}||_{Frob}^2 = \sum_{i}\sum_{ j }(\mathbf{D}_{ij}^\mathcal{T})^2 = \mathbf{tr}\mathbf{D}^\mathcal{T} (\mathbf{D}^\mathcal{T})^T
	\label{multi_class_wsd1_imp}
\end{equation}
where the element $\mathbf{D}_{ij}^\mathcal{T}$ in distance matrix $\mathbf{D}^\mathcal{T}$ is the $W_1$ distance between empirical class conditional distribution $p^\prime(\mathbf{X}_{\mathcal{T}}|y=i)$ and $p^\prime(\mathbf{X}_{\mathcal{T}}|y=j)$.

\subsection{Properties of 1-Wasserstein Distance}
The $W_1$ distance possesses some good properties which make it prevalent in a wide range of domain \cite{rubner2000earth}.
For the demand of feature selection, the main characteristics of Wasserstein distance which contribute a lot are the following:
\begin{itemize}
	\item As a metric: It's symmetric, non-negative, and satisfies the triangular inequality, metrics the weak convergence of probability measures.
	\item Robustness: It's insensitive to small perturbations of the distributions, and it can handle distributions with disjoint supports or different dimensions, moreover it could compare discrete and continuous distributions.
	\item Expressiveness: It takes into account the geometric structure of the underlying space and captures all modes of difference between distributions.
\end{itemize}

\subsubsection{Metric Structure}

The first thing to note is that the $W_1$ distance is a metric, so it has a good theoretical justification, consistency in feature selection.

The $W_1$ distance is a distance measure on the space of probability measures defined on the measurable space $\mathcal{X}$.
It is based on the comparison of integrals of the 1-Lipschitz function $f\in \mathcal{F}$ with respect to two probability measures $P$ and $Q$.
The distance $W_1$ has a metric structure, i.e. it satisfies the following properties for arbitrary probability measures $P,Q$ and $R$ on $\mathcal{X}$.
\begin{itemize}
\item Non-negativity: $W_1(P,Q)\ge0$ , and $ W_1(P, Q) = 0$ if and only if $P$ and $Q$ are the same.
\item Symmetry: $W_1(P, Q)$ = $W_1(Q, P)$.
\item Triangle inequality: $W_1(P,R)\le W_1(P,Q)+W_1(Q,R)$ .
\end{itemize}
The metric structure of the $W_1$ distance implies that it metrizes the weak convergence of probability measures, i.e. $W(P_n,P)\rightarrow 0$ if and only if $P_n$ weakly converges to $P$.
It guarantees the convergence $\int _{\mathcal{X}}fdP_{n}\rightarrow \int _{\mathcal{X}}fdP$ for all bounded continuous functions $f$ on $\mathcal{X}$, which means that $W_1$ has a consistent and unbiased estimator on finite samples of $P$ and $Q$.

More specifically, the non-negativity property means that $W_1$ distance is always positive or zero, and it is zero if and only if the two probability measures are equal. This means that $W_1$ distance can distinguish between different probability distributions, which guarantees the expressiveness in feature selection.
The symmetry property means that the $W_1$ distance is the same regardless of the order of the two probability measures, it implies that the $W_1$ distance does not depend on the direction or orientation of the comparison, it's fair and consistent for both sides.
The triangle inequality property of $W_1$ distance makes it applicable to define balls or neighborhoods around any probability measure, which are sets of probability measures that are within a certain distance from the central measure.
This also means that $W_1$ distance can be used to define convergence or continuity of sequences or functions of probability measures, notions that depend on how close or far apart the terms or values are in terms of the W1 distance.

\subsubsection{Robustness}

In practice, the dataset used to select features is often noisy or unbalanced, which poses challenges for feature selection criteria.
1-Wasserstein could be robust to measure the distance between empirical class conditional distributions, for it could compare distributions with disjoint supports or even different dimensions, insensitive to perturbations, and is consistent with translation and constant scaling.

Since the p-Wasserstein distance is defined in terms of the cost of moving mass between the two distributions and does not rely on pointwise comparison, it can compare distributions with disjoint supports.
Note that there are no assumptions about the dimension of the distributions, e.g., given $P,Q$ on $\chi_1,\chi_2$ respectively, the $\Gamma(P,Q)$ consists of couplings which is a joint distribution on $\chi_1 \times \chi_2$ with marginals $P$ and $Q$.
The optimal coupling in such a problem could be found by applying Brenier's theorem and solving Monge-Ampere equations.

Due to its metric nature, the p-Wasserstein distance $W_p$ is consistent with translation and constant scaling, which makes it robust to small perturbations of the distributions:
\begin{itemize}
	\item $\forall a\in R, W_p(aP,aQ)= |a| W_P(P,Q)$
	\item $\forall \mathbf{x}\in \mathcal{X}, W_p(P+\mathbf{x}, Q+\mathbf{x}) = W_p(P,Q)$
	\item $\forall \mathbf{x}\in \mathcal{X}, W_{p}(P+\mathbf{x},Q)=||\mathbf{x}||_{p}+W_{p}(P,Q)$
\end{itemize}

The 1-Wasserstein distance is more robust than p-Wasserstein distances when $p>1$ due to the norm used in the definition, meaning it's less insensitive to small perturbations of the probability measures.
Suppose the distribution of the noise is $O$, the empirical distribution of the mixing noise to the original $\tilde{P}_n$.
$$ \tilde{P}_n=P*O $$
where $*$ is the convolution operator.
When we estimate the p-Wasserstein distance between P and Q using the noisy samples, the estimator is $W_{p}(\tilde{P}_{n},Q)$.
According to the triangle inequality, we have 
$$ | W_{p}(\tilde{P}_{n},Q) - W_{p}(P,Q) | \le W_{p}(\tilde{P}_{n},P)$$
Using Jensen's Inequality, we can see that the 1-Wasserstein distance takes the smallest error when faced with noisy data for $p\ge 1$.

We must note that the standard p-Waterstein distance is sensitive to outliers, since all the mass in the original distribution should be transported to the target distribution. However, there are many works that focus on improving the robustness of the Wasserstein distance.
\cite{balaji2020robust} introduced a dual form of robust optimal transport optimization that is computationally efficient and suitable for deep learning applications.
\cite{mukherjee2021outlierrobust} derived an outlier robust optimal transport formulation by adding a total variation constraint, which allows ignorance of outliers from the original distribution and makes it compatible with existing computational optimal transport methods.

\subsubsection{Expressiveness}

The criterion based on $W_1$ distance is expressive, because it could detect all types of differences between distributions without assuming any parametric models or requiring overlapping supports, and it takes into account the underlying geometry of the space in which the distributions are defined.

By definition, the $W_1$ distance measures the minimum cost of transporting mass from one distribution to another, it does not rely on any pointwise comparison or alignment of the distributions, but only on the optimal transport map that minimizes the cost.
This means that it can detect the difference between distributions with different shapes, supports, modes, or dimensions.
Therefore, the criteria constructed based on 1-Wasserstein distance is zero if and only if all class conditional distributions of certain feature are equal, which means there is no discriminative information provided by certain feature in probability sense.

Moreover, due to the cost of moving mass in the optimal transport problem based on geodesic distance, the $W_1$ distance takes into account the geometry of the underlying space, e.g., if $P$ and $Q$ are degenerate at points $x,y\in \mathcal{X}$ respectively, the resulting Wasserstein distance is equal to the distance between $x$ and $y$.
This property allows the use of Wasserstein distances in problems that need to take geometric information into account \cite{rubner2000earth, arjovsky2017Wasserstein}, and also suggests the feasibility and effectiveness of feature selection criteria based on the 1-Wasserstein distance in such domains. 

\subsection{Estimation of Criteria based on Wasserstein Distance}

Since the feature selection criterion is a simple operation on the distance between the empirical class conditional distributions, its estimation is straightforward.
Denote the estimation or approximation of the $W_1$ distance by $\hat{W_1}$, we can obtain the approximate estimation of the distance matrix $\hat{\mathbf{D}}$ and the utility $\hat{U(\mathcal{T})}$ of the feature set $\mathcal{T}$ on this basis.
\begin{equation}
	\hat{U(\mathcal{T})} = ||\hat{\mathbf{D}}^\mathcal{T}||_{Frob}^2 = \sum_{i}\sum_{ j }(\hat{\mathbf{D}}_{ij}^\mathcal{T})^2 = \mathbf{tr}\hat{\mathbf{D}}^\mathcal{T} (\hat{\mathbf{D}}^\mathcal{T})^T
	\label{multi_class_wsd1_imp_appro}
\end{equation}
where the $\hat{\mathbf{D}}_{ij} = \hat{W_1}(p^{\prime} (\mathbf{X}_\mathcal{T} | y=c_i) , p^{\prime} (\mathbf{X}_\mathcal{T} | y=c_j))$ is the approximate estimation of $W_1$ distance between empirical class conditional distribution corresponding to feature set $\mathcal{T}$ and class $c_i$ and $c_j$.
Thus, we focus here the estimation of $W_1$ distance only.

The $W_1$ distance means the minimum cost of transforming one probability measure into another, it can be seen as the optimum of the associated optimal transport problem, so estimating the Wasserstein distance is actually solving the associated constrained optimization problem.
Only for some special cases, the optimal transport problem can be solved in closed form, such as one-dimensional distributions and Gaussian distributions, in which case we can estimate the Wasserstein distance directly.
For other cases, we need to approximate the Wasserstein distance by various methods, e.g., linear programming, kernel approximation, smooth approximation, random methods and generative approximation.
Here, we briefly introduce the exact computation on 1-dimensional distributions and the approximate estimation by the entropic regularization technique, for we utilize the former to construct a feature evaluation criterion and the latter to construct criteria for evaluating feature sets.

\subsubsection{Exact Computation}

Most of the time, numerical solutions are needed to compute the Wasserstein distance, only in some special cases there are efficient or even closed-form solvers.
For instance, when the dimensions of the probability measures are equal to one, or both measures are Gaussian, in such cases, the optimal transport problem could be solved in closed form.

If the probability measures are 1-dimensional, so that the underlying space is $\mathcal{X}=R$, there is a simple formula for the optimal map between the probability measures, i.e., mapping each quantile of one distribution to the corresponding quantile of the other.
Given empirical probability measure $P_m=\frac{1}{m}\sum_{i=1}^{m}\delta_{x_i}, Q_n=\frac{1}{n}\sum_{i=1}^{n}\delta_{y_j}$, supposing that samples are ordered in a same way, we can compute the p-Wasserstein distance on this two measures by
$$W_{p}(P_m , Q_n)= (\frac{1}{n}\sum _{i=1}^{n}|x_{i}-y_{i}|^{p}) ^ {\frac{1}{p}}$$
We can generalize this formula by the quantile function of probability measures
$$W_{p}(P, Q) = (||C_{P}^{-1}-C_{Q}^{-1}||_{L^p(\left[ 0,1 \right])}^{p})^{\frac{1}{p}} = (\int _{0}^{1}|C_{P}^{-1}(r)-C_{Q}^{-1}(r)|^{p}dr)^{\frac{1}{p}}, p\ge1$$
where the $C_{P}$ is the cumulative distribution function of $P$, $C_{P}^{-1}$ is the generalize quantile function of $P$.
$$C_{P}^{-1}(r)= \min \left\{ x \in R \cup \left\{ - \infty \right\} :C_{P}(x)\geq r \right\},\quad \forall r \in \left[ 0,1 \right].$$
We can simplify the computation formula when p=1
$$W_{1}(P, Q) = ||C_{P}-C_{Q}||_{L^1(\left[ 0,1 \right])} = \int _{R}|C_{P}(x)-C_{Q}(x)|dx.$$
The computational complexity of computing the Wasserstein distance on 1-dimensional distributions is $O(nlogn)$.
Thus it could be computationally friendly if we can evaluate the distance between high-dimensional distributions on the basis of Wasserstein distance in pairs of 1-dimensional distributions, which leads to the famous Sliced Wasserstein \cite{rabin2012wasserstein} and will appear in the feature selection algorithms introduced in this paper.

\subsubsection{Approximate Computation}

There are often motivations to approximately estimate the $W_1$ distance, for the associated optimal transport problem is often computationally intractable or impractical to solve it exactly.
In such a case, we need to estimate it approximately by using some methods that can reduce the computational complexity or simplify the problem structure, such as linear programming, kernel approximation, and random methods, etc.
Here we present some of them in a nutshell for the completeness and the neat of this paper, consider the case that computing $W_1$ distance on discrete probability measure $P_m=\frac{1}{m}\sum_{i=1}^{m}\delta_{x_i}, Q_n=\frac{1}{n}\sum_{i=1}^{n}\delta_{y_j}$.
Linear programming is a straightforward way to solve such optimal transport problem.
For instance, we can estimate the $W_1$ distance by a nonparametric introduced by \cite{sriperumbudur2012empirical}, and the estimator is on the basis of solving a linear program problem.
However, it's computationally expensive to solve this problem, the running time in the worst case could be $O(n^3logn)$ \cite{pele2009fast} for $P$ and $Q$ supported on the same number of points.
To relieve the burden of computation, there are many works focus on improving the efficiency to solve the linear program associated to the optimal transport problem.
Some of them attempt to develop specialized algorithms for special cases, while many people try to improve the solving algorithms \cite{sommerfeld2019optimal}.

Another way to approximately estimate the Wasserstein distance is to modify the associated optimal transport problem, most of them are on the basis of random methods or regularization techniques.
\cite{rabin2012wasserstein} proposed Sliced Wasserstein Distance by changing the original Wasserstein distance on high-dimensional probability measures to expectation over random projections of the distributions onto 1D subspaces, for the 1D Wasserstein distance can be computed exactly.
\cite{cuturi2013sinkhorn} modified the optimal transport problem by adding an entropic regularization term to the objective function, which leads to a strictly convex optimization problem and could be solved by Sinkhorn algorithm with complexity $O(n^2)$, several orders of magnitude less than the that of classical one \cite{panaretos2019statistical}.
We utilize this approximate estimator for multidimensional distributions, for it can be computed by simple matrix scaling algorithm, which makes it much faster than classical linear programming solvers and GPU friendly.
Moreover, it might be more stable and robust than the unregularized Wasserstein distance of optimal transport.

There are many algorithms and variants in the literature, we refer to \cite{peyre2019computational} for more details and comprehensive demonstration.

\subsection{Convergence}
Convergence of empirical estimators of Wasserstein distance on empirical measures are well studied topics in probability theory and statistics.
There are many results that vary depending on the dimension, the order, and the regularity of the underlying probability measures.
Here we present some of them for completeness.
To keep the neatness of article, we present literature works about convergence behavior in two cases: exact computation on empirical measures in 1-dimensional space, and the Sinkhorn algorithm on multi-dimensional empirical measures.

We first consider the $W_1$ distance on 1-dimensional probability measures. Given an arbitrary empirical probability measure $P_m=\frac{1}{m}\sum_{i=1}^{m}\delta_{x_i}$ in metric space, it's well known that if the underlying space is separable, with probability one, $P_m \rightarrow P$ weakly \cite{varadarajan1958convergence}.
If under the p-Wasserstein distance, with probability one, $W_p(P_m,P)\rightarrow 0$ for $p\ge1$ as $n\rightarrow \infty$.
Moreover, \cite{bobkov2019onedimensional} showed that the p-Wasserstein distance in the expected sense, $E(W_p(P_m,P))\rightarrow 0$ for $1\le p <\infty$ as $n\rightarrow \infty$.
They also pointed out that the worst rate of $E(W_p(P_m,P))$ on 1-dimensional distributions could be arbitrarily slow, and the best convergence rate $\frac{1}{\sqrt{n}}$ of $W_1$ distance on 1-dimensional empirical measures would not be reached unless the measure satisfies
$$
J_1(P)=\int_{-\infty}^{\infty} \sqrt{F_P(x)(1-F_P(x))} dx < \infty
$$
where $F_P$ is the distribution function of $P$.
Fortunately, we are often given samples with finite values so that the associated empirical measure satisfies this condition naturally, i.e., we could achieve the best rate in practice.

As is mentioned in Section \ref{convergence_of_ipms}, there are many works that focus on the convergence of $W_1$ distance to empirical measures.
\cite{sriperumbudur2009integral} showed that their nonparametric empirical estimator of $W_1(P_m,Q_n)$ is strongly consistent, they also gave the convergence rate of it.
Given a bounded, convex subset $M$ of $(R^d, ||\cdot||_s)$ with nonempty interior, probability measures $P$ and $Q$ are defined on $M$, then
\begin{equation}
	\begin{aligned}
|W(P_m,Q_n) - W(P,Q)| & = O_{P,Q}(r_m+r_n)\\
r_m & = \left\{\begin{array}{cl}
	m^{-1 / 2}, & d=1 \\
	m^{-1 / 2} \log m, & d=2 \\
	m^{-1 / d}, & d>2
\end{array}\right.
	\end{aligned}
\label{convergence_rate_of_WSD}
\end{equation}
note that if the $M$ is just a bounded subset of $(R^d, ||\cdot||_s)$, the convergence rate could be slower.

The convergence rate of the $W_1$ distance to empirical measures depends on the properties of the underlying space $M$ and the estimator, it would converge very slowly if the dimension of $M$ is large, if the estimator could not be free from the curse of dimensionality.
Moreover, there may not exist closed form solution for the corresponding optimal transport problem in high-dimensional space, we need to approximate the $W_1$ distance in such cases.
The Sinkhorn algorithm for entropically regularized optimal transport \cite{cuturi2013sinkhorn} is a prevalent method to approximately compute the Wasserstein distance, we evaluate the feature set $\mathcal{T}$ on the basis of this estimator.
The Sinkhorn distance induced by the modified optimal transport problem could converge to the Wasserstein distance as the regularization coefficient goes to zero, and could satisfy the conditions of metric if revised, besides, the Sinkhorn algorithm has a linear convergence \cite{franklin1989scaling}.

Therefore, the utility of the feature $f_t\in\mathcal{T}$ evaluated using the criteria constructed based on the $W_1$ distance is also concentrated and takes a good convergence rate.

\subsection{Feature Selection Algorithms}
\label{wsd_feature_selection_algorithms}

In this section, we describe methods for feature selection based on the proposed framework.

After defining the feature selection criteria, we describe algorithms that perform feature selection based on this dependency measure.
Let $\mathcal{S}$ be the full set of features, $\mathcal{T}$ be a subset of features $\mathcal{T}\subset \mathcal{S}$, we want to find $\mathcal{T}^\star$ such that the distance between the class conditional distribution restricted to it is maximized.
We can choose different metrics or divergences to measure the difference between distributions for different aspects we concern in certain task, here we use the $W_1$ distance as an example to show how to do feature selection with our framework.

Remember that solving the global feature selection problem is NP-hard, exhaustive search could obtain the optimal solution while taking the cost of exponential growth of computational cost, it's common to use some heuristic search strategy to approximate the optimum.
Moreover, we should choose the appropriate one among different feature selection strategies with respect to certain task, we can evaluate features individually and select the top-$m$, or we can build up a catalog of features in an incremental way (forward selection), or whether we would like to remove irrelevant features from a catalog (backward selection).
The top-$m$ strategy is computationally more efficient than sequential selection, at the cost of ignoring the interaction between features.
Forward selection requires less computational cost than backward elimination, but backward elimination generally yields a better feature subset (especially for nonlinear features), since the utility of the feature subset is evaluated in the context of all other features \cite{guyon2002gene}.

\subsubsection{Top-$m$ Using $W_1$ Distance (TWD) }
We first introduce a filter feature selection method constructed in a naive way based on the proposed framework.
We employ the $W_1$ distance between feature-wise class conditional distributions to construct evaluation criteria, and then sum up the utility of each feature as the utility of the feature subset $\mathcal{T}$
$$
U_{W_1} (\mathcal{T}) = \sum_{t=1}^{m} U_{W_1} (f_t), |\mathcal{T}| = m
$$
It's worth noting that we cannot obtain an exact estimate of the Wasserstein distance even though the class-dependent distribution of a given feature is a 1-D distribution, because there is no analytic expression for the class-dependent distribution.
All we can do is to compute the Wasserstein distance on empirical measures based on samples corresponding to certain feature and class.
In this case, each feature corresponds to a distance matrix, the entries of this matrix are the $W_1$ distance between corresponding class conditional probability density of certain feature, it contains all the statistical behavior of the feature among all classes in distributional discrepancy sense.
The importance of the feature here is the Frobenius norm of the corresponding distance matrix.
To obtain a feature subset, we use the simplest top-$m$ strategy, where $m$ features with the largest importance score are chosen as the result of feature selection.
This may be the most naive strategy, but it actually works in practice.

This algorithm has some advantages that make it feasible in different areas and stages.
It evaluates features individually according to meaningful criteria, which could provide insights for knowledge discovery.
It runs relatively fast, such that it could cooperate with other strategies or even methods to balance the effectiveness and computational cost.
The drawback of it is that it could not handle the redundancy in the feature set due to the lack of consideration of interaction among features.
However, we can reduce the redundancy in selected features by means of considering the similarity between distance matrices.
Since the distance matrix captures the entire discriminative behavior of features in all classes, we can study the redundancy based on the distance matrix.
It's feasible to study the relationship between features by measuring the similarity between the corresponding distance matrices.

\subsubsection{Forward Add-in Using $W_1$ Distance(FAWD)}
As mentioned above, we can account for the interaction between features in our framework by using a sequential feature selection procedure.

FAWD uses a sequential forward add-in selection strategy that starts with an empty set $\mathcal{T}_f$, evaluates all possible unselected features, and adds a feature to $\mathcal{T}_f$ each step to achieve the greatest utility as measured by our criteria \ref{multi_class_wsd1_imp}, until the size of $\mathcal{T}_f$ reaches the desired size $m$.
The $\mathcal{T}_f$ is the solution to \ref{feature_selection_problem}, and we can take the first $l, l\ge m$ elements from $\mathcal{T}_f$ if we need a smaller feature set.

The distance between the class conditional distributions is monotonically increasing in this procedure, but the gain from adding a feature $f_t$ does not mean the utility of the individual feature $U(f_t)$, because the interaction between $f_t$ and the selected features is taken into account.
Note that this procedure is nonparametric and model-free, thus preserving the generality of the data type and the relationship between data and labels.

Also, we can make a trade-off between runtime and effectiveness by controlling the number of features added in $\mathcal{T}_f$ at each step \cite{song2012feature}, that is, we can add a group of features at each step for faster selection.

\subsubsection{Backward Elimination Using $W_1$ Distance(BEWD)}
BEWD uses the opposite approach to FAWD, it utilizes sequential backward elimination strategy, starting from the full feature set $\mathcal{S}$ and eliminating at each step a feature which influence the distance among class conditional distributions the least.

This procedure generates a list of features $\mathcal{S}^c$ which containing the features eliminated from $\mathcal{S}$.
At each step, one feature from $\mathcal{S}$ is added to $\mathcal{S}^c$ which is not already in $\mathcal{S}^c$ yet, and we can pick the last $m$ elements from $\mathcal{S}^c$ as the solution to \ref{feature_selection_problem}.
Like the FAWD, we can eliminate a group of features from $\mathcal{S}$ at each step for efficiency.

\section{Experiments}
\label{Experiments}
In this section, we carry out experiments\footnote{github URL will come soon.} to analyze the proposed feature selection methods.
We implement them on real-world datasets and compare them with existing popular feature selection methods.

\subsection{Data Sets}

We have utilized benchmark datasets from various domains, including image data, voice data, sensor data, and biomedical data, to benchmark feature selection methods in the literature \cite{balin2019concrete, lemhadri2021lassonet}. These datasets can be classified into two categories based on their background and tasks: pattern recognition datasets (COIL-20, MNIST, Fashion-MNIST, and nMNIST-AWGN) and tabular datasets (Activity, ISOLET, and MICE). The former consists of images, while the latter consists of structural items. We have conducted experiments on these datasets to compare the effectiveness and versatility of our method with existing approaches. Table \ref{details_of_datasets} summarizes the characteristics of the datasets used in our experiments\footnote{We have resized the images in the COIL-20 dataset to 20*20 images using the same pre-processing step as \cite{lemhadri2021lassonet}, resulting in a total of 400 features.}.

\begin{itemize}
	\item COIL-20: The COIL-20 dataset is a collection of 1,440 grayscale images of 20 objects, each of which was captured in 72 different poses. This dataset has been widely used in computer vision research for object recognition and classification tasks.
	
	\item MNIST: The MNIST dataset is a standard benchmark for computer vision and machine learning, consisting of 60,000 handwritten digits for training and 10,000 for testing. It has been extensively used in research on deep learning, image recognition, and pattern recognition.
	
	\item Fashion-MNIST: The Fashion-MNIST dataset is a more challenging and realistic alternative to the original MNIST dataset, consisting of 70,000 images of fashion items. This dataset has been used to evaluate the performance of various machine learning algorithms for image classification tasks.
	
	\item nMNIST-AWGN: The nMNIST-AWGN dataset is an extension of the MNIST dataset created by adding white Gaussian noise to the original images to simulate noisy environments. This dataset has been used to evaluate the robustness of machine learning algorithms to noise in image classification tasks.
	
	\item Human Activity Recognition (Activity): The Human Activity Recognition (Activity) dataset contains numerical items collected by sensors in smartphones, each item corresponding to one of six poses of the user. This dataset has been used to evaluate the performance of various machine learning algorithms for activity recognition tasks.
	
	\item ISOLET: The ISOLET dataset is a collection of spoken letters of the English alphabet, where the cardinality of each sample is a feature extracted from voice data. This dataset has been used to evaluate the performance of various machine learning algorithms for speech recognition tasks.
	
	\item MICE: The MICE dataset is a collection of gene expression data of 77 proteins from 1080 mice with different genetic backgrounds and treatments. This dataset has been used to evaluate the performance of various machine learning algorithms for gene expression analysis.
\end{itemize}

\begin{table}[!h]
	\centering
	\begin{adjustbox}{max width=\textwidth,keepaspectratio} 
		\begin{tabular}{|c|c|c|c|c|c|c|c|}
			\hline
			Dataset & Activity & COIL-20 & ISOLET & MICE & MNIST & Fashion-MNIST & nMNIST-AWGN \\ \hline
			\#(Features) & 561 & 400 & 617 & 77 & 784 & 784 & 784 \\ \hline
			\#(Samples) & 5744 & 1440 & 7797 & 1080 & 70000 & 70000 & 70000 \\ \hline
			\#(Classes) & 6 & 20 & 26 & 8 & 10 & 10 & 10 \\ \hline
		\end{tabular}
	\end{adjustbox}
	\caption{The Details of Benchmark Datasets}
	\label{details_of_datasets}
\end{table}

To ensure a fair comparison, we divided each dataset into training and test sets using the same random state. However, to avoid memory overflow on our PC, we split the MNIST, Fashion-MNIST, and nMNIST-AWGN datasets into training and test sets with a ratio of 20-80. In contrast, we split the COIL-20, MICE, and ISOLET datasets into training and test sets with a 70-30 ratio because they have relatively few samples compared to the number of classes. In this way, we can ensure that each dataset is split in a way that is optimal for the characteristics of that dataset.

\subsection{Methodology}
We compare our method with several filter feature selection methods that were constructed based on statistical or information theoretic criteria. including JMI \cite{yang1999data}, CFS \cite{hall1999feature}, Fisher Score \cite{duda2000pattern},  Trace-ratio \cite{nie2008trace}, MRMR\cite{hanchuanpeng2005feature}, CMIM \cite{fleuret2004fast},  DISR \cite{meyer2008informationtheoretic}, UDFS \cite{yang2011l2}.
We also compare popular embedded approaches,  such as HSIC-Lasso \cite{yamada2014highdimensional},  LassoNet \cite{lemhadri2021lassonet}. 
To ensure a fair comparison, we used the scikit-feature  \cite{li2018feature} implementation of each method wherever available. 
We implement HSIC-Lasso using the pyHSIC package \cite{yamada2014highdimensional} \cite{climente-gonzalez2019block}, and download the source code from the official repository\footnote{https://github.com/lasso-net/lassonet} to implement LassoNet. 
We set the hyperparameters of compared methods at default. By doing so, we can ensure that our method is compared to other methods in a way that is optimal for the specific dataset’s characteristics.

We use different methods to select feature sets, and feed them into independent classifiers to explore their effectiveness in ensuring fairness. We compare each feature selection method by selecting a varying number of features, and measure the accuracy obtained by downstream classifiers when fed with these features. To quantify the performance of feature selection methods, we use the accuracy obtained as a metric. We report the average results of these methods from 10 runs on different training sets. To ensure fairness, we randomly split the dataset and ensure that all methods work on the same data in each run.

We conducted our experiments in Python on a desktop with an Intel Core i7-12700KF CPU @4.6GHz and 32GB physical RAM.

\begin{figure}[]
	\centering
	\begin{subfigure}[b]{0.49\textwidth}
		\includegraphics[width=\textwidth]{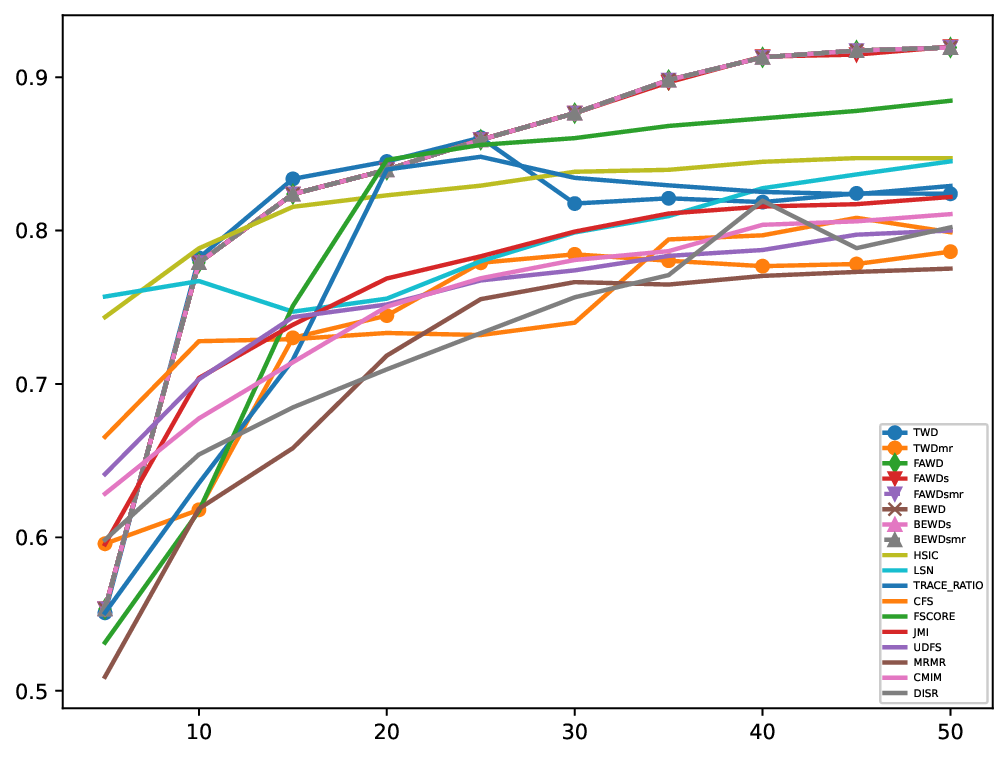}
		\caption{Activity}
		\label{Activity_xgb_acc}    
	\end{subfigure}
	\begin{subfigure}[b]{0.49\textwidth}
		\includegraphics[width=\textwidth]{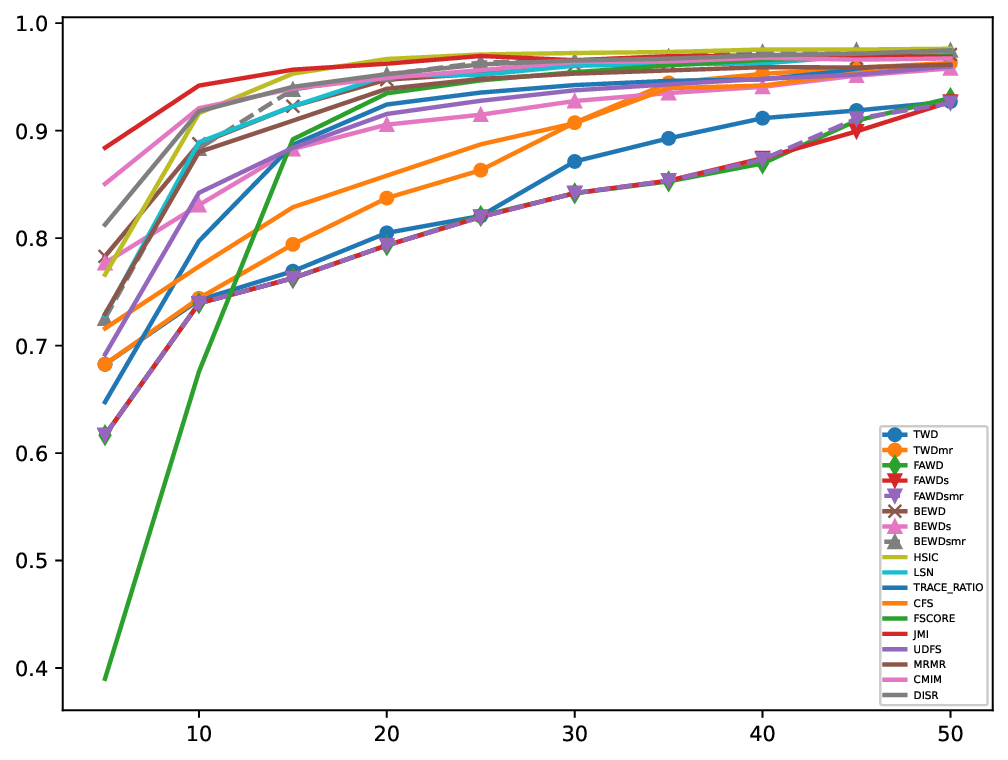}
		\caption{COIL-20}
		\label{COIL-20_xgb_acc}    
	\end{subfigure}
	\begin{subfigure}[b]{0.49\textwidth}
		\includegraphics[width=\textwidth]{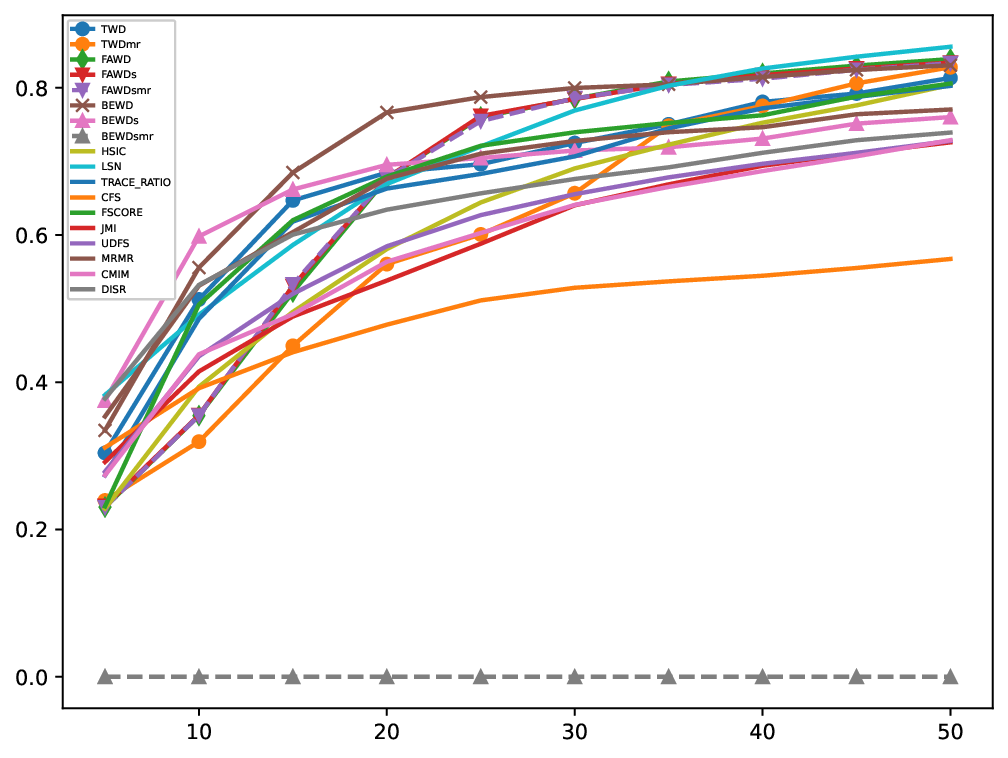}
		\caption{ISOLET}
		\label{ISOLET_xgb_acc}    
	\end{subfigure}
	\begin{subfigure}[b]{0.49\textwidth}
		\includegraphics[width=\textwidth]{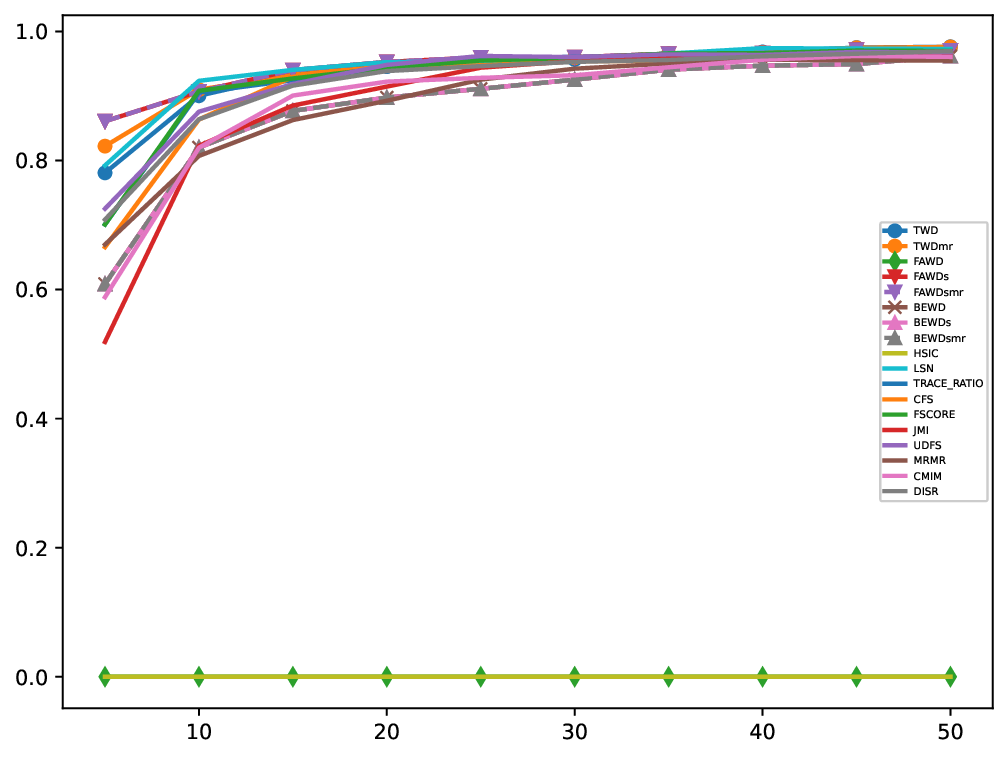}
		\caption{MICE}
		\label{MICE_xgb_acc}    
	\end{subfigure}
	\begin{subfigure}[b]{0.49\textwidth}
		\includegraphics[width=\textwidth]{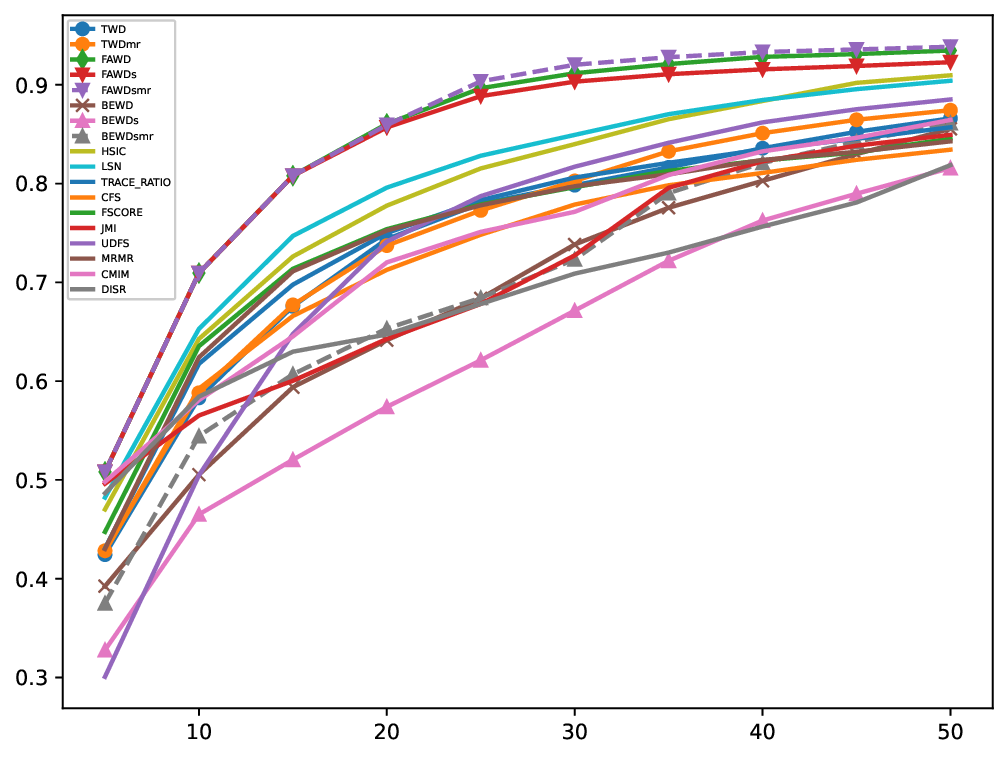}
		\caption{MNIST}
		\label{MNIST_xgb_acc}    
	\end{subfigure}
	\begin{subfigure}[b]{0.49\textwidth}
		\includegraphics[width=\textwidth]{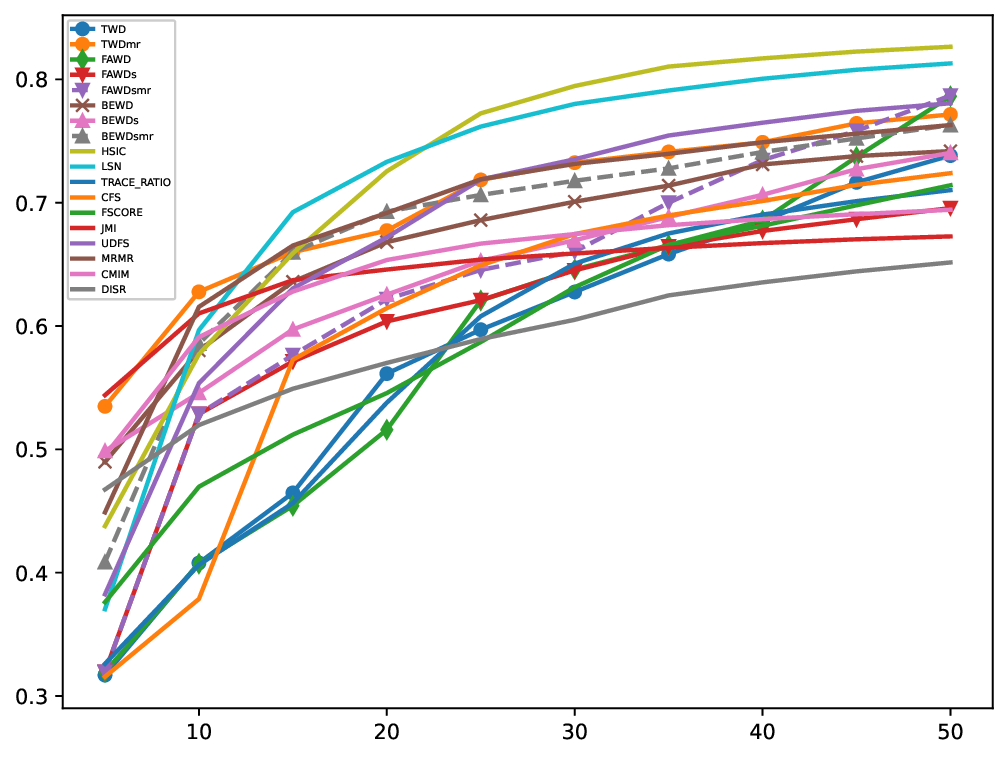}
		\caption{Fashion-MNIST}
		\label{MNIST-Fashion_xgb_acc}    
	\end{subfigure}
	\caption{Classification Accuracy.}
	\label{xgb_acc}
\end{figure}

\subsection{Classification accuracy}
We use the accuracy of classification to evaluate the value of features in supervised learning. This metric is also commonly used to compare the performance of feature selection methods. We evaluate different methods by varying the number of features they select. For the Mice and COIL-20 datasets, which are small in size, we test the accuracy of feature sets with five different sizes ranging from 5 to 25. For the other datasets, we test the feature sets with sizes ranging from 10 to 50.

To implement the classifier, we choose XGBoost \cite{chen2016xgboost} for its advantages in performance, interpretation, and ease of control. We set the same hyperparameters for all feature sets, set the number of trees to 50, and fix the random state of the classifiers to 0 to ensure a fair comparison.

Our proposed framework performs well on all benchmark datasets from different domains. Compared with similar filter methods, our framework achieves the best or nearly the best accuracy when the number of features ranges from 5 to 50. Compared with embedded methods such as LSN and HSIC, our framework leads to the highest or nearly the highest accuracy on all datasets except Fashion-MNIST. The experimental results also show that our method is effective on datasets with small sample sizes, such as COIL-20 and MICE. In general, greedy search outperforms the top-$m$ strategy, and backward elimination is more applicable than forward add-in, although both can achieve good results.

We run our experiments in Python on a desktop with an Intel Core i7-12700KF CPU @4.6GHz and 32GB physical RAM.



\subsection{Stablility}

Feature selection methods can be evaluated for stability from two perspectives: robustness to sample perturbations and resilience to noise in the data. We conducted experiments on both aspects to demonstrate the stability of our method.

To investigate the stability of our method in the face of instance-wise perturbation, we ran feature selection methods against each other 10 times on benchmark datasets. In each run, we randomly split the datasets with a predefined random state to ensure that each method faced the same data, and the random state numbers were chosen randomly.

\begin{figure}[]
	\centering
	\begin{subfigure}[b]{0.49\textwidth}
		\includegraphics[width=\textwidth]{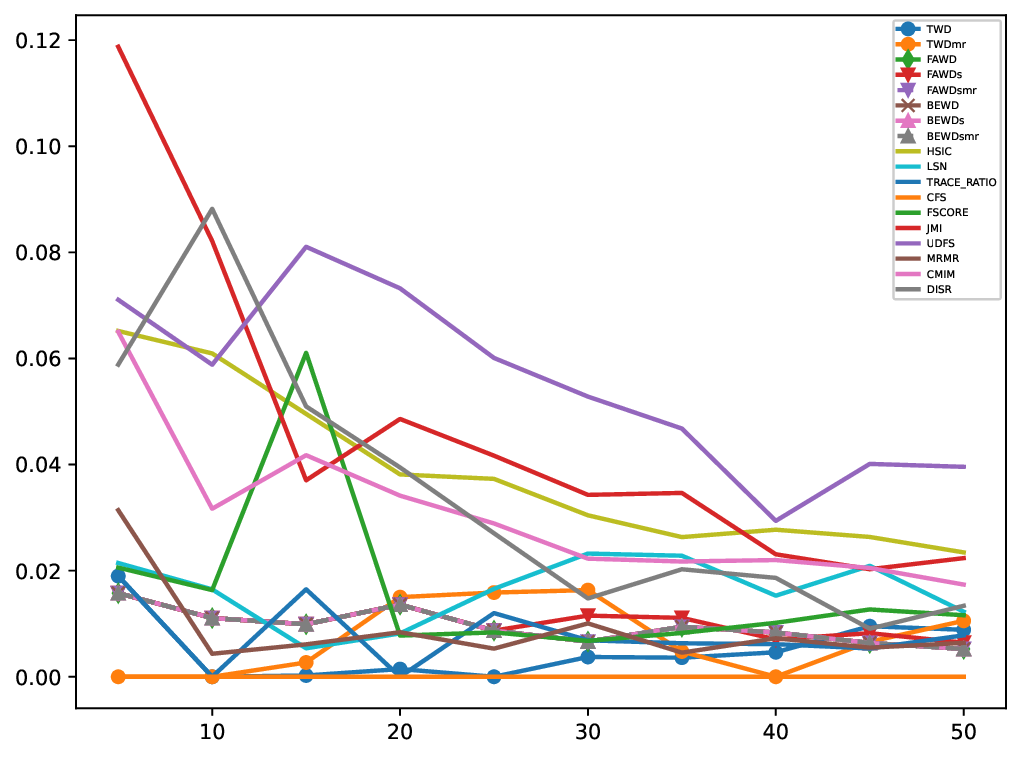}
		\caption{Activity}
		\label{Activity_xgb_rsd}    
	\end{subfigure}
	\begin{subfigure}[b]{0.49\textwidth}
		\includegraphics[width=\textwidth]{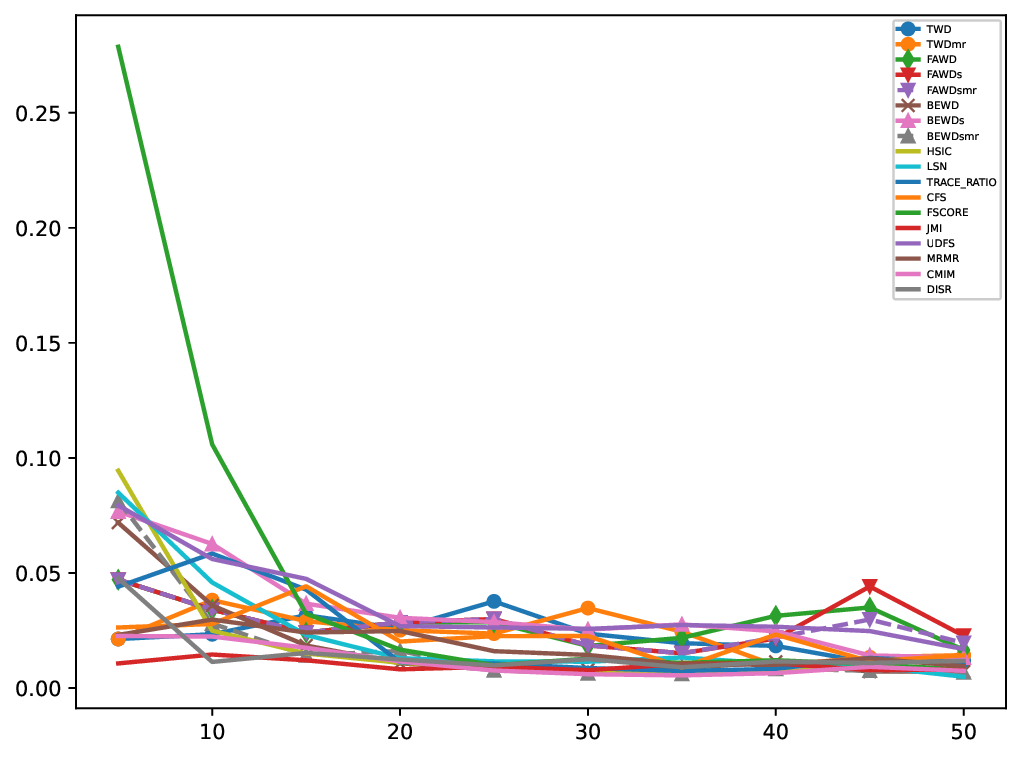}
		\caption{COIL-20}
		\label{COIL-20_xgb_rsd}    
	\end{subfigure}
	\begin{subfigure}[b]{0.49\textwidth}
		\includegraphics[width=\textwidth]{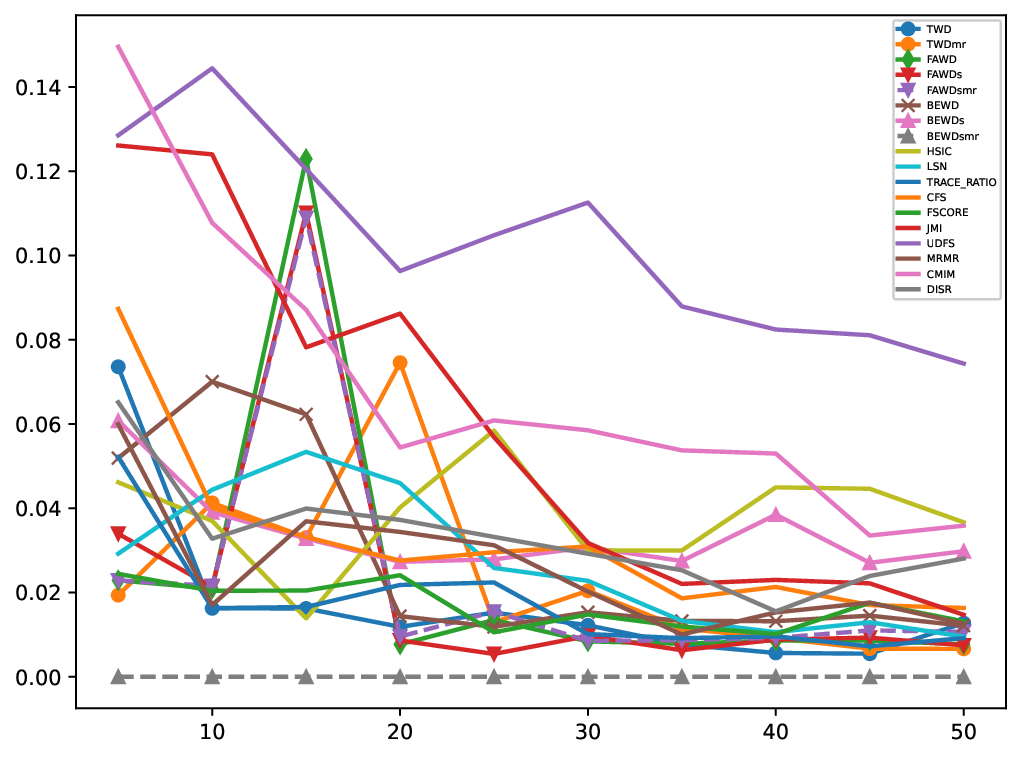}
		\caption{ISOLET}
		\label{ISOLET_xgb_rsd}    
	\end{subfigure}
	\begin{subfigure}[b]{0.49\textwidth}
		\includegraphics[width=\textwidth]{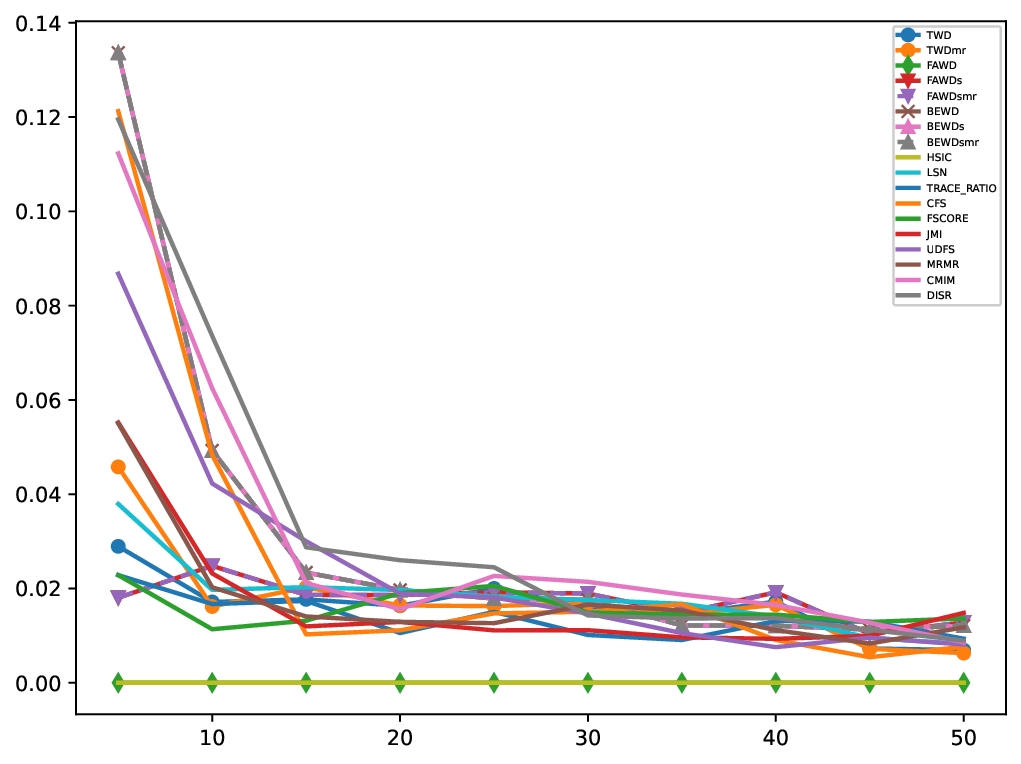}
		\caption{MICE}
		\label{MICE_xgb_rsd}    
	\end{subfigure}
	\begin{subfigure}[b]{0.49\textwidth}
		\includegraphics[width=\textwidth]{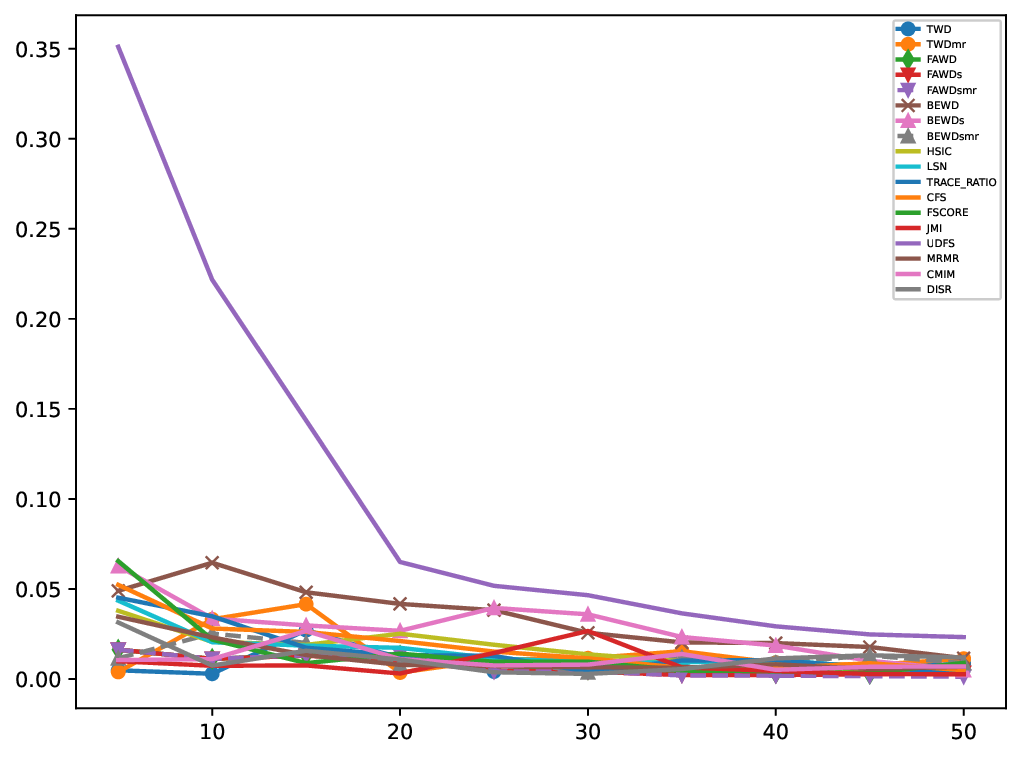}
		\caption{MNIST}
		\label{MNIST_xgb_rsd}    
	\end{subfigure}
	\begin{subfigure}[b]{0.49\textwidth}
		\includegraphics[width=\textwidth]{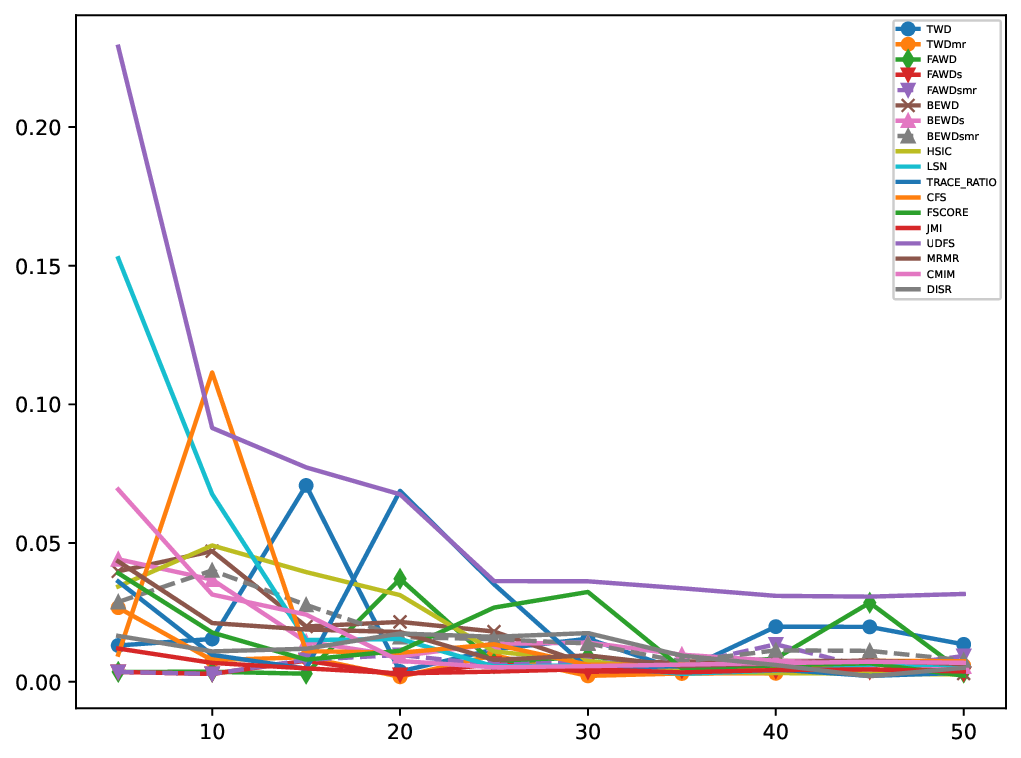}
		\caption{Fashion-MNIST}
		\label{MNIST-Fashion_rsd}    
	\end{subfigure}
	\caption{Relative standard deviation.}
	\label{xgb_rsd}
\end{figure}

We used the relative standard deviation (RSD) of accuracy as a tool to quantify the stability of feature selection methods. The RSD is the ratio of the standard error of the accuracy to the accuracy.
$$
RSD = \frac{Std(Accuracy) }{Accuracy}
$$
As shown in figure \ref{xgb_rsd}, our methods shows the least RSD on all benchmark datasets, means that they have excellent stability, although not throughout the whole process of varying the number of feature.
The results show that our framework could perform stably when facing instance-wise perturbations.
Moreover , we find that the TWD is more stable than FAWD and BEWD, which suggests that sequential search strategies may be more vulnerable than top-$m$.

We also investigate the robustness of our proposed techniques by testing them on noisy data. The noise in the data requires a larger number of samples to obtain a reliable estimate of a given accuracy level, which poses some difficulties for feature selection methods. To verify the stability of our techniques, we use the nMNIST-AWGN dataset, which is created by adding white Gaussian noise to the images from MNIST. The empirical results show that the techniques derived from our framework perform admirably in noisy scenarios. FAWD achieves the highest performance throughout the feature number variation process, and TWD also performs near the top. BEWD has a moderate performance, which can be attributed to the fact that BEWD initially includes some noisy features and also retains some of them during the backward elimination process. This can lead to a worse performance than TWD, which does not consider the interactions between features.

\begin{figure}[]
	\centering
	\begin{subfigure}[b]{0.99\textwidth}
		\includegraphics[width=\textwidth]{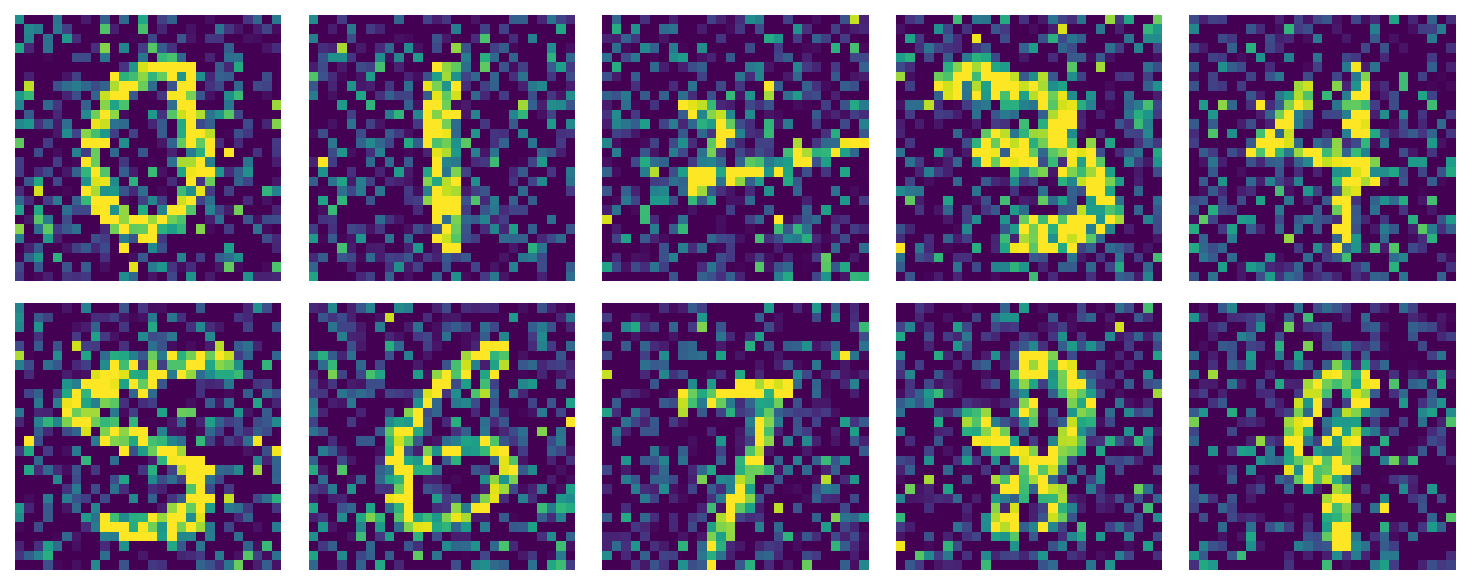}
		\caption{}
		\label{nmnist_awgn_samples}    
	\end{subfigure}
	\begin{subfigure}[b]{0.49\textwidth}
		\includegraphics[width=\textwidth]{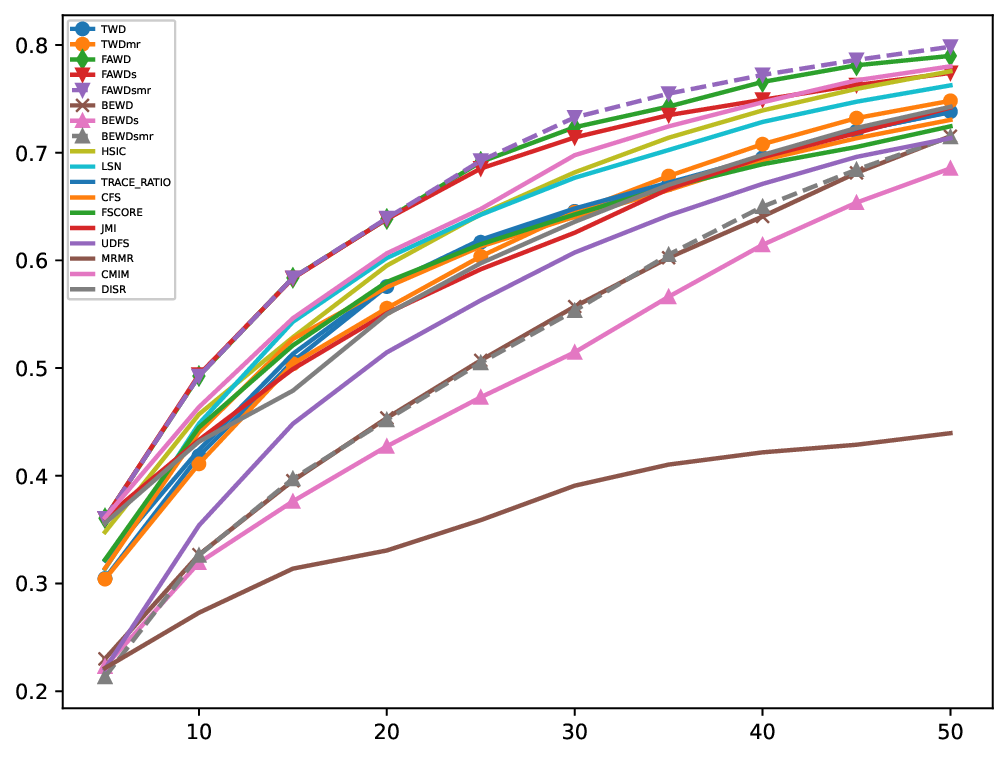}
		\caption{}
		\label{nmnist_awgn_acc}    
	\end{subfigure}
	\begin{subfigure}[b]{0.49\textwidth}
	\includegraphics[width=\textwidth]{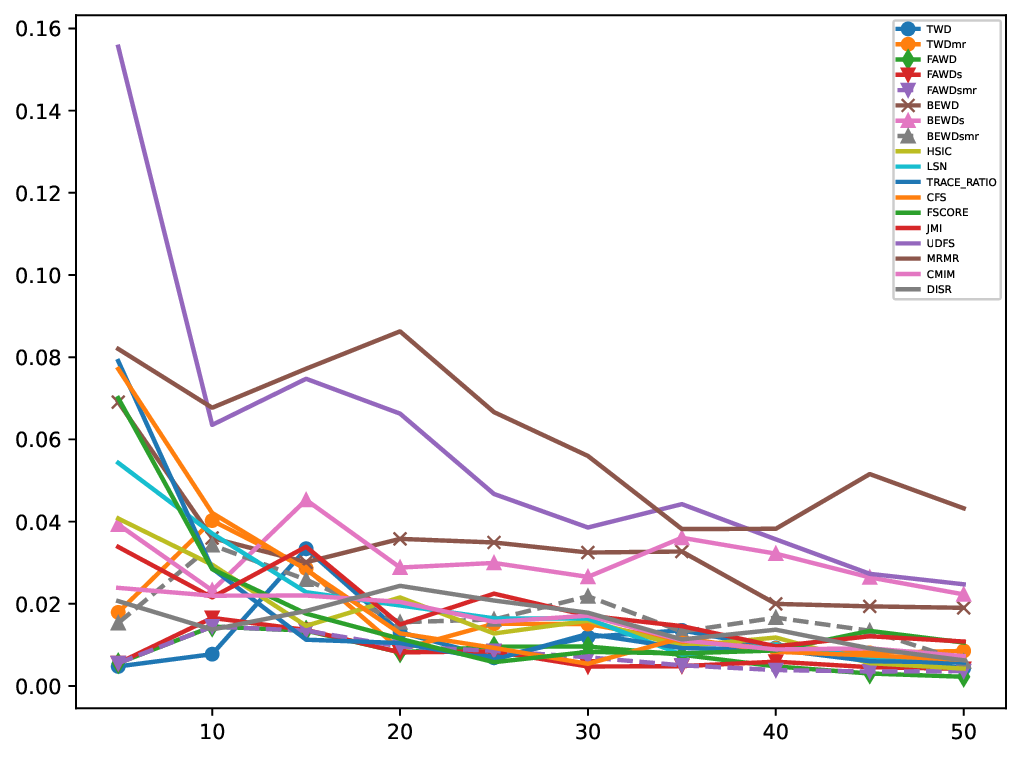}
	\caption{}
		\label{nmnist_awgn_rsd}    
\end{subfigure}
	\caption{Experiments on nMNIST-AWGN. Figure (a) shows some randomly selected images from different classes of the nMNIST-AWGN dataset. Figure (b) shows the average accuracy achieved by the feature subsets obtained from feature selection on the nMNIST-AWGN dataset. Figure (c) shows the RSD values for different feature selection methods.}
	\label{xgb_rsd_nmnist}
\end{figure}

\section{Conclusion}
\label{Conclusion}
In this paper, we introduce a novel approach to feature selection based on the distance between class conditional distributions. Our method uses integral probability metrics (IPMs) to evaluate features. Our approach differs from conventional methods in that it directly explores the discriminative information in terms of feature distributions, which is intrinsic to supervised classification problems. We systematically present the theoretical justifications and estimation methods of IPMs that are useful for feature selection, and demonstrate that our approach is suitable for a variety of tasks and strategies. We develop a variant of our approach based on the 1-Wasserstein distance, evaluate its feasibility for feature selection, present exact and approximate estimation methods, and analyze their convergence properties. We conduct experiments on real-world benchmark datasets from different domains and compare them with other methods. The experimental results show that our approach is suitable for a variety of domains and tasks and has excellent robustness.

Python code and documentation for the proposed methods are available at \textbf{Github url coming soon.}.

\subsection*{Acknowledgments}

\newpage

\bibliography{feature_selection_bibt.bib}

\bibliographystyle{nips-no-url}
\newpage

\end{document}